\documentclass[11pt]{article}

\pdfoutput=1 
\usepackage[a4paper,hscale=0.725,vscale=0.78]{geometry}
\usepackage[font=it,labelfont=bf,margin=10pt]{caption}

\RequirePackage{amsthm,thmtools}
\newtheorem{theorem}{Theorem}
\theoremstyle{definition}
\newtheorem{definition}[theorem]{Definition}

\newtheorem{safety property}[safetyprop]{Safety Property}

\usepackage{amsmath}
\usepackage{amsfonts}
\usepackage{url}

\usepackage{newtxtext}

\usepackage[linkcolor=black,bookmarksopen,bookmarksopenlevel=1,breaklinks=true]{hyperref}

\DeclareMathOperator*{\argmax}{argmax}

\RequirePackage{subcaption}

\def\bif{\text{\bf if}\,}
\def\bthen{\,\text{\mbox{\bf then}}\,}
\def\belse{\,\text{\mbox{\bf else}}\,}

\newcommand*{\Util}{\mathcal{U}}

\def\EE{\mathbb{E}}

\def\TypeOf#1{\textbf{Typeof}_{#1}}

\usepackage[table]{xcolor}
\usepackage{dm-colors}

\definecolor{darkgreen}{RGB}{0,64,0}
\definecolor{lightgreen}{RGB}{128,255,128}

\RequirePackage{tikz}
\usetikzlibrary{arrows.meta}
\tikzset{>={latex}}

\usepackage[decisionutilitycolor,nonumber]{influence-diagrams}

\usepackage[utf8]{inputenc}
\tikzset{
  vallabel/.style = {rectangle, fill=yellow,minimum height=0.45cm, minimum width=0.7cm},
}

\tikzset{
  ann/.style = {draw=none, rectangle, minimum size=0mm,inner sep=0mm}
}
\tikzset{
  learn/.style = {color=blue,rectangle, minimum size=0mm,inner sep=1mm}
}
\tikzset{
  wide/.style = {line width=0.6pt}
}

\colorlet{player3-color}{dmteal100}
\tikzset{
  player3/.style = {fill=player3-color}
}

\newenvironment{smallitemize}{
\vspace{-.5ex}
\begin{itemize} \setlength{\itemsep}{0ex} 
}{
\end{itemize}
\vspace{-1ex}
}

\newenvironment{smallenumerate}{
\vspace{-1.5ex}
\begin{enumerate} \setlength{\itemsep}{0ex} 
}{
\end{enumerate}
\vspace{-1.5ex}
}

\def\percent{\%}
\def\ozero{o\raisebox{-1mm}{$_0$}}
\DeclareMathOperator*{\concat}{\text{+\hspace*{-0.7ex}+}}

\usepackage{calc}

\newlength{\agentdeftw}
\newenvironment{agentdef}[1]{
\vspace{.5ex}
~~{\bf #1} \hfill
\setlength{\agentdeftw}{\textwidth-\widthof{~~{\bf #1}~}}
\begin{minipage}[t]{\agentdeftw}
}{
\end{minipage}
\vspace{.5ex}
}

\def\ishift{0mm}
\def\iannotshift{0mm}
\def\nodestretch{1.8cm}
\def\ashift{0.3cm}
\def\basiciagentelements{
      \node (S0) [] {$X_0$};
      \node (A0) [right = \ashift of S0,yshift=1.3cm] {$A_0$};
      \node (S0i) [above = 1mm of S0, ann] {$x_0$};
      \node (A0i) [above = 1mm of A0, ann] {$\pi$};

      \node (S1) [right = \nodestretch of S0] {$X_1$};
      \node (A1) [right = \ashift of S1,yshift=1.3cm] {$A_1$};
      \node (S1i) [above = 1mm of S1, ann] {$[X]$};
      \node (A1i) [above = 1mm of A1, ann] {$\pi$};

      \node (S2) [right = \nodestretch of S1] {$X_2$};
      \node (A2) [right = \ashift of S2,yshift=1.3cm] {$A_2$};
      \node (S2i) [above = 1mm of S2, ann] {$[X]$};
      \node (A2i) [above = 1mm of A2, ann] {$\pi$};

      \node (S3) [right = \nodestretch of S2] {$X_3$};
      \node (S3i) [above = 1mm of S3, ann] {$[X]$};

      \node (I0) [below = \auxvspace of S0,xshift=\ishift] {$I_0$};
      \node (I1) [right = \nodestretch of I0] {$I_1$};
      \node (I2) [right = \nodestretch of I1] {$I_2$};
      \node (I3) [right = \nodestretch of I2] {$I_3$};
      \node (I0i) [above = 1mm of I0, ann,xshift=-0.5mm] {$i_0$};
      \node (I1i) [above = 1mm of I1, ann,xshift=\iannotshift] {$[I]$};
      \node (I2i) [above = 1mm of I2, ann,xshift=\iannotshift] {$[I]$};
      \node (I3i) [above = 1mm of I3, ann,xshift=\iannotshift] {$[I]$};
      
      \edge{S0,I0,A0}{S1,I1};	
      \edge{S1,I1,A1}{S2,I2};
      \edge{S2,I2,A2}{S3,I3};
     
      \edge{S0,I0}{A0};
      \edge{S1,I1}{A1};
      \edge{S2,I2}{A2};

}

\usepackage{tocloft}
\begin{document}
%%%!4
%%%!2

\title{\bf Counterfactual Planning in AGI Systems}
\author{{\bf Koen Holtman}\\[.5ex]
{\normalsize Eindhoven, The Netherlands}\\
%{\normalsize January 2021}\\
{\normalsize \tt Koen.Holtman@ieee.org}}
%\date{}
\date{January 2021}
\maketitle
\begin{abstract}
We present counterfactual planning as a design approach for creating a
range of safety mechanisms that can be applied in hypothetical future
AI systems which have Artificial General Intelligence.

The key step in counterfactual planning is to use an AGI machine
learning system to construct a counterfactual world model, designed to
be different from the real world the system is in.  A counterfactual
planning agent determines the action that best maximizes expected
utility in this counterfactual planning world, and then performs the
same action in the real world.

We use counterfactual planning to construct an AGI agent emergency
stop button, and a safety interlock that will automatically stop the
agent before it undergoes an intelligence explosion.  We also
construct an agent with an input terminal that can be used by humans
to iteratively improve the agent's reward function, where the
incentive for the agent to manipulate this improvement process is
suppressed.  As an example of counterfactual planning in a non-agent
AGI system, we construct a counterfactual oracle.

As a design approach, counterfactual planning is built around the use
of a graphical notation for defining mathematical counterfactuals.
This two-diagram notation also provides a compact and readable language for
reasoning about the complex types of self-referencing and indirect
representation which are typically present inside machine learning
agents.
\end{abstract}

%{\blue Note: Editorial and meta comments not intended for inclusion in
%any final version are typeset in {\bf blue} and sometimes {\bf \red
%red} text.}

\setcounter{tocdepth}{1}
\tableofcontents
\setcounter{tocdepth}{3} % for hyperref

\section{Introduction}%%%!2

Artificial General Intelligence (AGI) systems are hypothetical future
machine reasoning systems that can match or exceed the capabilities of
humans in general problem solving.  While it is still unclear if AGI
systems could ever be built, we can already study AGI related risks
and potential safety mechanisms
\cite{bostrom2014superintelligence,russell2019human,everitt2018agi}.

In this paper, we introduce counterfactual planning as a design
approach for creating a range of AGI safety mechanisms.
Counterfactual planning is built around a graphical modeling system
that provides a specific vantage point on the internal construction of
machine learning based agents. This vantage point was designed to make
certain safety problems and solutions more tractable.

An AI agent is an autonomous system which is programmed to use its
sensors and actuators to achieve specific goals.  A well-known risk
in using AI agents is that the agent might mispredict the results of
its own actions, causing it to take actions that produce a
disaster.  The main risk driver we consider here is different. It is
the risk that an inaccurate or incomplete specification of the agent
goals produces a disaster. 

Any AGI agent goal specification created by humans will likely be
somewhat inaccurate, no matter whether it is created by hand-coding or
by machine learning from selected examples
\cite{p1,hadfield2019incomplete}.  If one gives an even slightly
under-specified goal to a very powerful autonomous system, there is a
risk that the system may end up perfectly achieving this goal, while
also producing several unexpected and very harmful side effects.  This
motivates research into AGI emergency stop buttons, interlocks which
can limit the power of the agent, and safe ways to update the goal
while the agent runs.

\subsection{Use of natural and mathematical language}

When writing about AGI systems, one can use either natural language,
mathematical notation, or a combination of both.  A natural
language-only text has the advantage of being accessible to a
larger audience.  Books like {\it Superintelligence}
\cite{bostrom2014superintelligence} and {\it Human Compatible} \cite
{russell2019human} avoid the use of mathematical notation in the main
text, while making a clear an convincing case for the existence of
specific existential risks from AGI, even though these risks are
currently difficult to quantify.

However, natural language has several shortcomings when it is used to
explore and define specific technical solutions for managing AGI
risks.  One particular problem is that it lacks the means to
accurately express the complex types of self-referencing and indirect
representation that can be present inside online machine learning
agents and their safety components.  To solve this problem, we
introduce a compact graphical notation.  This notation unambiguously
represents these internal details by using two diagrams: a {\it
learning world diagram} and a {\it planning world diagram}.

\subsection{AGI safety as a policy problem}%%%!

Long-term AGI safety is not just a technical problem, but also a
policy problem.  While technical progress on safety can sometimes be
made by leveraging a type of mathematics that is only accessible to
handful of specialists, policy progress typically requires the use of
more accessible language.  Policy discussions can move faster, and
produce better and more equitable outcomes, when the description of a
proposal and its limitations can be made more accessible to all
stakeholder groups.

One specific aim of this work is to develop a comprehensive vocabulary
for describing certain AGI safety solutions, a vocabulary that is as
accessible as possible.  However, the vocabulary we develop has too much
mathematical notation to be accessible to all members of any possible
stakeholder group.  So the underlying assumption is that each
stakeholder group will have access to a certain basic level of
technical expertise.

At several points in the text, we have also included comments that aim
to explain and demystify the vocabulary and concerns of some specific
AGI related sub-fields in mathematics, technology, and philosophy.

\subsection{Related work that uses counterfactuals}%%%!

In the general AI/ML literature that is concerned with improving
system performance, counterfactual planning has been used to improve
performance in several application domains.  See for example
\cite{zinkevich2008regret} and \cite{bottou13a}, where the latter is
notable because it includes an accessible discussion about computing
confidence intervals for counterfactual projections.

In the AI safety/alignment literature, there are several system
designs which add counterfactual terms to the agent's reward function.
Examples are
\cite{corra, holtmanitr} in the AGI-specific safety literature,
and \cite{turner2020conservative,krakovna2018penalizing} which
consider both AI and AGI level systems.

In the literature on encoding specific human values into machine
reasoning systems, counterfactuals have been used to encode
non-discriminatory fairness towards individuals, for example in
\cite{kusner2017counterfactual}, and also other human moral principles
in \cite{pereira2017agent}.

\subsection{Structure of this paper}%%%!

Sections 2 -- 4 introduce the main elements of our graphical notation
and modeling system.  Readers already familiar with Causal Influence
Diagram (CID) notation, as it is used to define agents, will be able
to skim or skip most of this material.

Sections 5 -- 7 specify three example counterfactual planning agents.
These are used in the remaining sections to illustrate further aspects
of counterfactual planning.  Starting from section 6.2, it should be
possible for all readers to skim or skip sections, or to read sections
in a different order.

\section{Graphical Models and Mathematical Counterfactuals}%%%!2
\label{sec:models}

The standard work which defines mathematical counterfactuals is the
book {\it Causality} by Judea Pearl \cite{pearlcausality}.  This book
mainly targets an audience of applied statisticians, for example those
in the medical field, and its style of presentation is not very
accessible to a more general technical audience.

Pearl is also mainly concerned with the use of causal models as {\it
theories about the real world} which can guide the interpretation of
statistical data.  Much of the discussion in {\it Causality} is
about questions of statistical epistemology and decision making.  In
this text, we will use causal models to construct {\it agent
specifications}, not theories about the world.  When we clarify
issues of epistemology here, they tend to be different issues.

The debate among philosophers about the validity of Pearl's
statistical epistemology is still ongoing, as is usual for such
philosophical debates.  In the AGI community, where the epistemology
of machine learning is a frequent topic of discussion, this has
perhaps made the status of mathematical counterfactuals as useful and
well-defined mathematical tools more precarious than it should be.

Because of these considerations, we have written this section to avoid
any direct reference to Pearl's definitions and explanations in
\cite{pearlcausality}, even though at a deeper mathematical level, we
define the same system of causal models and counterfactuals.

\subsection{Graphical world models}%%%!

A {\it world model} is a mathematical model of a particular world.
This can be our real world, or an imaginary world.  To make a
mathematical model into a model of a particular world, we need to
specify how some of the variables in the model relate to observable
phenomena in that world.

We introduce our graphical notation for building world models by
creating an example graphical model of a game world.  In the game
world, a simple game of dice is being played.  The player throws a
green die and a red die, and then computes their score by adding the
two numbers thrown.

We create the graphical game world model in thee steps:
\begin{enumerate}
\item We introduce three random variables
and relate them to observations we can make when the game is played
once in the game world.  The variable $X$ represents the observed
number of the green die, $Y$ is the red die, and $S$ is the score.
\item We draw the diagram in figure \ref{diaggame}.
\begin{figure}[h]
\centering
    \begin{tikzpicture}[
      node distance=0.7cm,
      every node/.style={
        draw, circle, minimum size=0.7cm, inner sep=0.5mm}]

      \node (X) [] {$X$};
      \node (Y) [below = 0.8cm of X] {$Y$};
      \node (S) [right = 0.8cm of X,yshift=-0.6cm] {$S$};
      \edge {X,Y} {S};

      \node (Xi) [above = 1mm of X, ann] {$[D]$};
      \node (Yi) [above = 1mm of Y, ann] {$[D]$};
      \node (Si) [above = 1mm of S, ann] {$sum$};

    \end{tikzpicture}
  \caption{Graphical model of the game of dice in the game world.}
  \label{diaggame}
\end{figure}
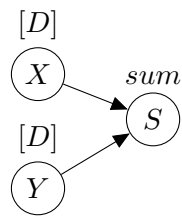
\item We define the two functions that appear in the {\it annotations}
above the nodes in the diagram:
\begin{equation*}
\begin{array}{l}
D(d) =
(\bif  d \in \{1,2,3,4,5,6\} \bthen 1/6 \belse 0 )~,\\[1ex]
sum(a,b)=a+b~.\\
\end{array}
\end{equation*}
\end{enumerate}

\subsubsection{Informal interpretation of the graphical model}

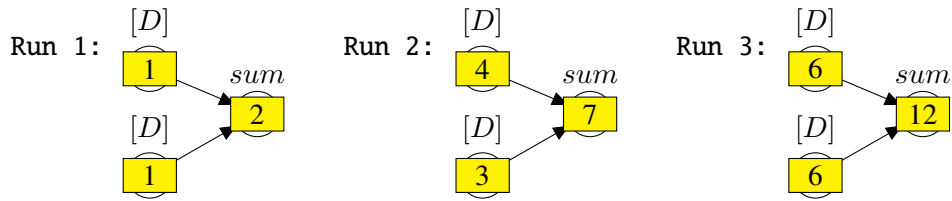
\begin{figure}[h]
\centering
\def\simrun#1#2#3#4{
    \begin{tikzpicture}[
      node distance=0.7cm,
      every node/.style={
        draw, circle, minimum size=0.6cm, inner sep=0.5mm}]

      \node (X) [] {$X$};
      \node (Y) [below = 0.8cm of X] {$Y$};
      \node (S) [right = 0.8cm of X,yshift=-0.6cm] {$S$};
      \edge {X,Y} {S};

      \node (Xi) [above = 1mm of X, ann] {$[D]$};
      \node (Yi) [above = 1mm of Y, ann] {$[D]$};
      \node (Si) [above = 1mm of S, ann] {$sum$};

     \node (name) [left=3mm of X, yshift=3mm,ann] {\tt Run #1:};
     \node (Xval) at (X.center) [vallabel] {#2};
     \node (Yval) at (Y.center) [vallabel] {#3};
     \node (Sval) at (S.center) [vallabel] {#4};
    \end{tikzpicture}}
\simrun{1}{1}{1}{2}~~~~~~
\simrun{2}{4}{3}{7}~~~~~~
\simrun{3}{6}{6}{12}
\caption{Using the graphical model as a canvas to display three
different simulator runs of the game world.}
  \label{sampleruns}
\end{figure}

We can read the above graphical model as a description of how we might
build a game world simulator, a computer program that generates random
examples of game play.  To compute one run of the game, the simulator
would traverse the diagram, writing an appropriate observed value into
each node, as determined by the function written above the
node. Figure \ref{sampleruns} shows three possible simulator runs.

\label{secsampleruns}
We can interpret the mathematical expression $P(S=12)$ as being the
exact probability that the next simulator run puts the number 12 into
node $S$.  This interpretation of $P(\cdots)$ expressions can be very
useful when reasoning informally about certain mathematical properties
of the graphical models.

The similarity between what happens in figure \ref{sampleruns} and
what happens in a spreadsheet calculation is not entirely
coincidental. Spreadsheets can be used to create models and
simulations without having to write a full computer program from
scratch.

\subsubsection{Formal interpretation of the graphical model}

In section \ref{formaldefs}, we will define the exact mathematical
meaning of drawing diagrams like figure \ref{diaggame}. The
definitions will treat the drawing as a Bayesian network, decorated
with three annotations written above the network nodes.  As an example
of how these definitions work, drawing the diagram in figure
\ref{diaggame} is equivalent to writing down the four equations
below, and declaring that these equations are mathematical sentences
with the truth value of `true'.
\begin{equation*}
\begin{array}{l}
%\forall_{x,y,s} ~
P(X=x,Y=y,S=s) = P(x=X)P(Y=y)P(S=s|X=x,Y=y)\\
P(X=x)=D(x)\\
P(Y=y)=D(y)\\
P(S=s|X=x,Y=y) = (\bif s = sum(x,y) \bthen 1 \belse 0)
\end{array}
\end{equation*}
The first equation above is produced by drawing the Bayesian network
graph, the other three are produced by adding the annotations.

To readers unfamiliar with Bayesian networks, the above equations may
look somewhat impenetrable at first sight. The key to interpreting
them is to note that the three right hand side terms of the first
equation appear on the left hand side in the next equations.
The equations therefore allow us to mechanically compute the
exact numerical value of $P(X=x,Y=y,S=s)$ for any $x$, $y$, and $s$,
by making substitutions until every $P$ operator is gone.  We can
compute that $P(X=1, Y=1, S=12)=0$.  We can compute that $P(S=12)=1/36$
by using that $P(S=12) = \sum_{x,y} P(X=x,Y=y,S=12)$.

A mathematical model can be used as a {\it theory} about a world, but
it can also be used as a {\it specification} of how certain entities
in that world are supposed to behave.  If the model is a theory of the
game world, and we observe the outcome $X=1, Y=1, S=12$, then this
observation falsifies the theory.  But if the model is a specification
of the game, then the same observation implies that the player is
doing it wrong.

\subsection{Counterfactuals}%%%!

We now show how mathematical counterfactuals can be defined using
graphical models.  The process is as follows.  We start by drawing a
first diagram $f$, and declare that this $f$ is the world model of a
{\it factual} world.  This factual world may be the real world, but
also an imaginary world, or the world inside a simulator.  We then draw
a second diagram $c$ by taking $f$ and making some modifications.  We
then posit that this $c$ defines a {\it counterfactual world}.  The
{\it counterfactual random variables} defined by $c$ then represent
observations we can make in this counterfactual world.

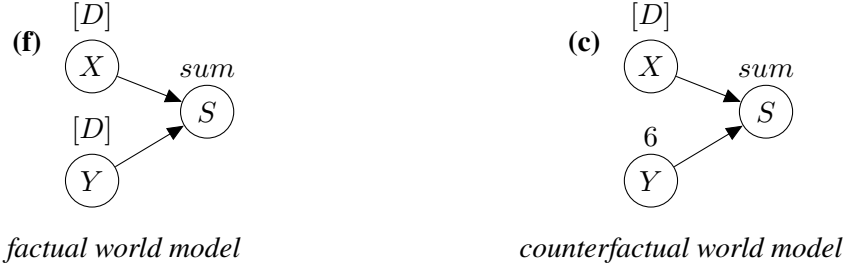
\begin{figure}[h]
  \begin{subfigure}{0.45\textwidth}
  \centering
    \begin{tikzpicture}[
      node distance=0.7cm,
      every node/.style={
        draw, circle, minimum size=0.7cm, inner sep=0.5mm}]

      \node (X) [] {$X$};
      \node (Y) [below = 0.8cm of X] {$Y$};
      \node (S) [right = 0.8cm of X,yshift=-0.6cm] {$S$};
      \edge {X,Y} {S};

      \node (Xi) [above = 1mm of X, ann] {$[D]$};
      \node (Yi) [above = 1mm of Y, ann] {$[D]$};
      \node (Si) [above = 1mm of S, ann] {$sum$};
      
      \node (name) [left=3mm of X, yshift=3mm,ann] {\bf(f)};
    \end{tikzpicture}\\[1ex]
    {\it factual world model}
    %\label{diagbn}
  \end{subfigure}
~~~
\begin{subfigure}{0.45\textwidth}
  \centering
    \begin{tikzpicture}[
      node distance=0.7cm,
      every node/.style={
        draw, circle, minimum size=0.7cm, inner sep=0.5mm}]

      \node (X) [] {$X$};
      \node (Y) [below = 0.8cm of X] {$Y$};
      \node (S) [right = 0.8cm of X,yshift=-0.6cm] {$S$};
      \edge {X,Y} {S};

      \node (Xi) [above = 1mm of X, ann] {$[D]$};
      \node (Yi) [above = 1mm of Y, ann] {$6$};
      \node (Si) [above = 1mm of S, ann] {$sum$};
      
      \node (name) [left=3mm of X, yshift=3mm,ann] {\bf(c)};
      
    \end{tikzpicture}\\[1ex] {\it counterfactual world model}
    \end{subfigure}
\caption{Example construction of a counterfactual
    world, with the model $c$ on the right defining three counterfactual
    random variables $X_c$,
    $Y_c$, and $S_c$.}  \label{constructcf}
\end{figure}

Figure \ref{constructcf} shows an example of the procedure, where we
construct a counterfactual game world in which the red die has the
number 6 on all sides.

We name diagrams by putting a label in the upper left hand corner.  In
figure \ref{constructcf}, the two labels {\bf (f)} and {\bf (c)}
introduce the names $f$ and $c$.  We will use the name in the label
for both the diagram, the implied world model, and the implied world.
So figure \ref{constructcf} constructs the counterfactual game world
$c$.

To keep the random variables defined by the above two diagrams apart,
we use the notation convention that a diagram named $c$ defines random
variables that all have the subscript $c$.  Diagram $c$ above
defines the random variables $X_c$, $Y_c$, and $S_c$. This convention
allows us to write expressions like $P(S_c > S_f) = 5/6$ without
ambiguity.

\subsection{Example model of an agent and its environment}

\begin{figure}[h]
  \centering
    \begin{tikzpicture}[
      node distance=0.7cm,
      every node/.style={
        draw, circle, minimum size=0.8cm, inner sep=0.5mm}]

      \node (S0) [] {$S_0$};
      \node (A0) [right = 0.5cm of S0,yshift=1cm] {$A_0$};
      \node (S0i) [above = 1mm of S0, ann] {$s_0$};
      \node (A0i) [above = 1mm of A0, ann] {$\pi$};

      \node (S1) [right = 1.8cm of S0] {$S_1$};
      \node (A1) [right = 0.5cm of S1,yshift=1cm] {$A_1$};
      \node (S1i) [above = 1mm of S1, ann] {$[S]$};
      \node (A1i) [above = 1mm of A1, ann] {$\pi$};

      \node (S2) [right = 1.8cm of S1] {$S_2$};
      \node (A2) [right = 0.5cm of S2,yshift=1cm] {$A_2$};
      \node (S2i) [above = 1mm of S2, ann] {$[S]$};
      \node (A2i) [above = 1mm of A2, ann] {$\pi$};

      \node (S3) [right = 1.8cm of S2] {$S_3$};
      \node (S3i) [above = 1mm of S3, ann] {$[S]$};

      \node (etc) [right = 0.6cm of S3,yshift=0.5cm,ann] {\bf ...};

      \edge{S0,A0}{S1};
      \edge{S1,A1}{S2};
      \edge{S2,A2}{S3};
     
      \edge{S0}{A0};
      \edge{S1}{A1};
      \edge{S2}{A2};
     
      \node (name) [left=15mm of A0, yshift=2mm,ann] {{\bf (d)}};
    \end{tikzpicture}
  \caption{Example diagram $d$ modeling an agent and its environment.}
  \label{diagd}
\end{figure}
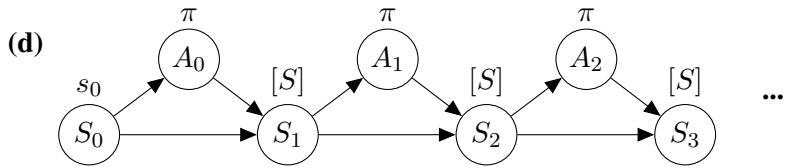

Diagram $d$ in figure \ref{diagd} models a basic MDP-style agent and
its environment.  The agent takes actions $A_t$ chosen by the policy
$\pi$, with actions affecting the subsequent states $S_{t+1}$ of the
agent's environment.  The environment state is $s_0$ initially, and
state transitions are driven by the probability density function $S$.

We interpret the annotations above the nodes in the diagram as {\it
model input parameters}. The model $d$ has the three input parameters
$\pi$, $s_0$, and $S$.  By writing exactly the same parameter above a
whole time series of nodes, we are in fact adding significant
constraints to the behavior of both the agent and the agent
environment in the model.  These constraints apply even if we specify
nothing further about $\pi$ and $S$.

We use the convention that the physical realizations of the agent's
sensors and actuators are modeled inside the environment states $S_t$.
This means that we can interpret the arrows to the $A_t$ nodes as
sensor signals which flow into the agent's compute core, and the arrows
emerging from the $A_t$ nodes as actuator command signals which flow
out.

\subsection{Formal definitions}%%%!
\label{formaldefs}

We now present fully formal definitions for the graphical
language and notation conventions introduced above.  The main reason
for including these is that we want to remove any possible ambiguity
from the agent definitions further below.

\subsubsection{Diagrams}

\begin{definition}[Diagram]
A diagram is a drawing that depicts a {\it graph}, which must be a
directed acyclic graph, by drawing nodes connected by arrows.  A {\it
node name}, starting with an uppercase letter, must be drawn inside
each node.  A node may also have an {\it annotation} drawn above it.
Drawings may use the notation `$\cdots$' to depict repeating
structures in the graph and its annotations.
\end{definition}

\subsubsection{Random variables and the $P$ notation}

We use random variables to represent observables in worlds.  We rely on
probability theory (see appendix \ref{randomvardef}) as the branch of
mathematics that defines truth values for expressions containing random
variables inside $P(\cdots)$ and $\EE(\cdots)$ operators.
Many texts use the convention that $P(s|x,y)$ is a
shorthand for $P(S=s|X=x,Y=y)$.  We avoid using this shorthand here,
partly to make the definitions below less cryptic, but also because it
tends to get typographically awkward when the random variables have
subscripted names.

\begin{definition}[Naming and subscripting of random variables]
When the graph drawn by a diagram with label {\bf (d)} has a node
named $X$ or $X_i$, then there exists a random variable named $X_d$ or
$X_{i,d}$ associated with that node. To avoid any ambiguity, we use
a comma to separate the two parts of the subscript in $X_{i,d}$.
\end{definition}

\subsubsection{Equations produced by drawing a diagram}

Before defining the equations produced by drawing a diagram, we 
define some auxiliary notation.

\begin{definition}[Parent notation $\Pa$ and $\pa$]
Let $X$ be the name of a graph node in diagram $d$, and let
$P_{1},\cdots,P_{n}$ be the list of names of all parent nodes of $X$,
all nodes which have an outgoing arrow into $X$.  The order in which
these parents appear in the list $P_{1},\cdots,P_{n}$ is determined by
considering each incoming arrow of $X$ in a clockwise order, starting
from the 6-o-clock position.  With this, $\Pa_{X,d}$ is the list of
random variable names $P_{1,d},\cdots,P_{n,d}$, and $\pa_{X,d}$ is the
list of lowercase variables names we get by converting the list
$P_{1},\cdots,P_{n}$ to lowercase.
\end{definition}

As an example, with figure \ref{diagd} above, $\Pa_{S_2,d}$ is the list
$S_{1,d},A_{1,d}$, and $\pa_{S_2,d}$ is the list $s_1,a_1$.

\begin{definition}[Bayesian model equation produced by drawing a diagram]
When we draw a diagram $d$ representing a graph with the nodes named
$X_1,\cdots X_n$, this is equivalent to stating that the following
equation is true:
\begin{equation*}
\begin{array}{l}
P(X_{1,d}=x_1,\cdots,X_{n,d}=x_n)=\\[1ex]
~~~~~~~P(X_{1,d}=x_1|\Pa_{X_1,d}=\pa_{X_1,d})
\cdot~ \ldots ~\cdot
P(X_{n,d}=x_n|\Pa_{X_n,d}=\pa_{X_n,d})
\end{array}
\end{equation*}
\end{definition}

\begin{definition}[Equation produced by adding an annotation]
When we draw an annotation above a node $X$ in a diagram $d$, then:
\begin{smallenumerate}
\item If the node has no parents and the annotation is a variable or
constant $v$, this is equivalent to stating that the following
equation is true:\\[1ex]
\hspace*{10ex}
$P(X_d=x)=(\bif x=v \bthen 1 \belse 0)$
\item If the node has parents and the annotation is a function $f$,
this states\\[1ex]
\hspace*{10ex}
$P(X_d=x | \Pa_{X,d}=\pa_{X,d})=(\bif x=f(\pa_{X,d})\bthen 1 \belse 0)$
\item If the node has parents and the annotation is $[F]$,
this states\\[1ex]
\hspace*{10ex}
$P(X_d=x | \Pa_{X,d}=\pa_{X,d})=F(x,\pa_{X,d})$\\[1ex]
where we require that the function $F$ satisfies
$\forall_{\pa_{X,d}} ( \sum_x F(x,\pa_{X,d}) = 1 )$.
\end{smallenumerate}
\end{definition}

\subsection{Differences with other notation conventions}%%%!

%Our use of Bayesian network node annotations to directly name model
%parameters is seems to be new, or at least it is uncommon.
%A more
%common style of Bayesian network annotation draws a full function
%table next to each node.

The {\bf do} notation is Pearl's most well-known device for defining
counterfactuals in a compact way.  We do not use this notation here,
because it is not well suited for defining the complex counterfactual
worlds we are interested in.

%We define such worlds by drawing
%diagrams instead.  In some cases, instead of drawing a new diagram, we
%just name it and then use natural language to describe how it should
%be drawn.

Pearl also defines a less well known notation in \cite{pearlcausality},
where subscripts are used to construct and label counterfactual random
variables.  This notation is different from the subscripting
conventions used here.

Many texts use the convention of introducing a model by writing down a
tuple like $(S,s_0,A,P,R,\gamma)$ which names all model parameters.
We do not use this convention here.  We introduce every model by
drawing a diagram, and name model parameters by drawing annotations in
the diagram.  This approach keeps several definitions in this text
much more compact, as we avoid having to translate back and forth
continuously between a graphical model representation and a
tuple-based representation.

\section{Causal Influence Diagrams}%%%!2

Influence diagrams \cite{howard2005influence} provide a graphical
notation for depicting utility-maximizing decision making processes.
In this paper we will use {\it Causal influence diagrams} (CIDs)
\cite{everitt2021causal}, a specific version of influence diagram notation
which has recently been proposed \cite{Everitt2019-2} for modeling and
comparing AGI safety frameworks.

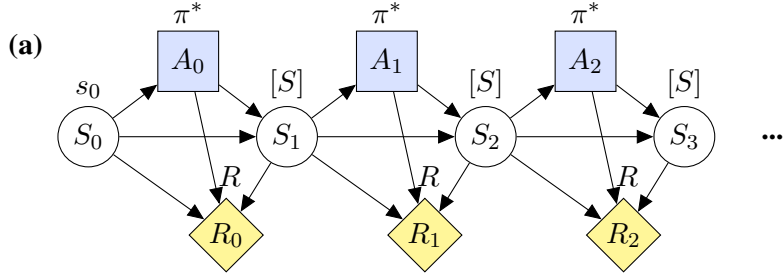
\begin{figure}[h]
  \centering \begin{tikzpicture}[ node distance=0.7cm, every
    node/.style={ draw, circle, minimum size=0.8cm, inner sep=0.5mm}]

% this is rouughly same as earlier diagram
      \node (S0) [] {$S_0$};
      \node (A0) [right = 0.5cm of S0,yshift=1cm,decision] {$A_0$};
      \node (S0i) [above = 1mm of S0, ann] {$s_0$};
      \node (A0i) [above = 1mm of A0, ann] {$\pi^*$};

      \node (S1) [right = 1.8cm of S0] {$S_1$};
      \node (A1) [right = 0.5cm of S1,yshift=1cm,decision] {$A_1$};
      \node (S1i) [above = 1mm of S1, ann] {$[S]$};
      \node (A1i) [above = 1mm of A1, ann] {$\pi^*$};

      \node (S2) [right = 1.8cm of S1] {$S_2$};
      \node (A2) [right = 0.5cm of S2,yshift=1cm,decision] {$A_2$};
      \node (S2i) [above = 1mm of S2, ann] {$[S]$};
      \node (A2i) [above = 1mm of A2, ann] {$\pi^*$};

      \node (S3) [right = 1.8cm of S2] {$S_3$};
      \node (S3i) [above = 1mm of S3, ann] {$[S]$};

      \node (etc) [right = 0.6cm of S3,yshift=0.0cm,ann] {\bf ...};

      \edge{S0,A0}{S1};
      \edge{S1,A1}{S2};
      \edge{S2,A2}{S3};
     
      \edge{S0}{A0};
      \edge{S1}{A1};
      \edge{S2}{A2};
    
      \node (name) [left=15mm of A0, yshift=2mm,ann] {\bf (a)};

      \node(R0) [below= 14mm of A0,xshift=5mm,utility] {$R_0$};
      \node(R0i) [above = 1.5mm of R0, ann] {$~R$};
      \edge{S0,S1,A0}{R0};

      \node(R1) [below= 14mm of A1,xshift=5mm,utility] {$R_1$};
      \node(R1i) [above = 1.5mm of R1, ann] {$~R$};
      \edge{S1,S2,A1}{R1};

      \node(R2) [below= 14mm of A2,xshift=5mm,utility] {$R_2$};
      \node(R2i) [above = 1.5mm of R2, ann] {$~R$};
      \edge{S2,S3,A2}{R2};

    \end{tikzpicture}
  \caption{Example causal influence diagram.  The diagram has
  diamond shaped {\em utility nodes} which define the value $\Util_a$,
  and square {\em decision nodes} which define $\pi^*$.  Optionally, colors
  can be used to highlight the structure of the diagram.
  }
  \label{diagoptpol}
\end{figure}

An example causal influence diagram is in figure \ref{diagoptpol}.
Our formal definitions of causal influence diagram notation extend the
definitions in \cite{everitt2021causal}: we also define multi-action
diagrams, and we add a time discount factor $\gamma$.

\subsection{Utility nodes defining expected utility}

If some nodes in a diagram are drawn with diamond shapes, these are
called {\it utility nodes}.  The {\it expected utility} of the diagram
is then defined as follows.

\begin{definition}[Expected utility $\Util_a$ of a diagram $a$]
We define $\Util_a$ for two cases:
\begin{smallenumerate}
\item If there is only one utility node $X$ in $a$, then $\Util_a =
\EE(X_a)$.
\item If there are multiple utility nodes $R_t$ in $a$, with integer
subscripts running from $l$ to $h$, then\\[-1em]
\begin{equation*}
\Util_a = \EE (~\sum_{t=l}^h~\gamma^t R_{t,a}~)
\end{equation*}
where $\gamma$ is a time discount factor, $0 < \gamma \leq 1$, which
can be read as an extra model parameter.
When $h=\infty$, we generally need $\gamma <1$ in order for $\Util_a$
to be well-defined.  
\end{smallenumerate}
\end{definition}

\subsection{Decision nodes defining the optimal policy}

When we draw some nodes in a diagram as squares, these are called {\it
decision nodes}. The purpose of drawing decision nodes is to define
the {\it optimal policy} which maximizes the expected utility of the
diagram.  We require that the same model parameter, the policy
function $\pi^*$ in the case of figure \ref{diagoptpol}, is present as
an annotation above all decision nodes.

\begin{definition}[Optimal policy $\pi^*$ defined by a diagram $a$]
A diagram $a$ with some utility and decision modes, where a function
$\pi^*$ is written above all decision nodes, defines this $\pi^*$ in
two steps.
\begin{smallenumerate}
\item First, draw a helper diagram $b$ by drawing a copy of
diagram $a$, except that every decision node has been drawn as a round
node, and every $\pi^*$ has been replaced by a fresh function name,
say $\pi'$.
%(With fresh, we mean that $\pi$ is not used
%elsewhere in the diagram or in other equations surrounding it.)
%
\item Then, $\pi^*$ is defined by 
$\pi^* = \argmax_{\pi'}~\Util_{b}$, where the $\argmax_{\pi'}$ operator always
deterministically returns the same function if there are several
candidates that maximize its argument.
\end{smallenumerate}
\end{definition}

\subsubsection{Approximately optimal policies}

In a real life agent implementation, the exact computation of the
optimal policy $\pi^*$ is usually intractable.  Only an approximately
optimal policy $\pi^+$ can be computed within reasonable time.  We
model this case as follows.

\begin{definition}[Approximately optimal policy $\pi^+$ defined by a
diagram $a$] A diagram $a$ where an optimal policy function $\pi^*$ is
written above all decision nodes also defines an approximately optimal
policy function $\pi^+$ by constructing the same helper diagram $b$
as above and then defining $\pi^+= \mathcal{A}(b)$, where the
function $\mathcal{A}$ processes the diagram $b$ and its model
parameter values to construct a policy $\pi'$ that does a reasonable
job at maximizing the value of $\Util_{b}$.
\end{definition}

%When $\mathcal{A}$ gets lucky, the approximately optimal policy $\pi^+$ that it
%constructs may end up being the same function as $\pi^*$.

To keep the presentation more compact, we will only use the optimal
policy symbol $\pi^*$ in the agent definitions below.

%But in any
%case, the definition could be rewritten using $\pi^+$ to define an
%equivalent approximately optimal policy agent.  This replacement should of
%course always trigger a consideration of the risk that the
%$\mathcal{A}$ chosen will produce a less perfect approximation of the
%intended safety properties.

\section{Online Machine Learning}%%%!2
\label{sec:machinelearning}

We now model {\it online machine learning agents}, agents that
continuously learn while they take actions.  These agents are also
often called {\it reinforcement learners}, see section
\ref{sec:rl} for a discussion which relates our modeling system to
reinforcement learning concepts and terminology.

We model online machine learning agents by drawing two diagrams, one
for a {\it learning world} and one for a {\it planning world}, and by
writing down an {\it agent definition}.
This two-diagram modeling approach departs from the approach in
\cite{Everitt2019-2,Everitt2019-3,everitt2021causal}, where only a single
CID is used to model an entire agent.  By using two diagrams instead
of one, we can graphically represent details
which remain hidden from view when using only a single CID.

\subsection{Learning world}

\begin{figure}[h]
  \centering
    \begin{tikzpicture}[
      node distance=0.7cm,
      every node/.style={
        draw, circle, minimum size=0.8cm, inner sep=0.5mm}]

% this is roughly same as earlier diagram
      \node (S0) [] {$S_0$};
      \node (A0) [right = 0.3cm of S0,yshift=1.3cm] {$A_0$};
      \node (S0i) [above = 1mm of S0, ann] {$s_0$};
      \node (A0i) [above = 1mm of A0, ann] {$\pi$};

      \node (S1) [right = 1.8cm of S0] {$S_1$};
      \node (A1) [right = 0.3cm of S1,yshift=1.3cm] {$A_1$};
      \node (S1i) [above = 1mm of S1, ann] {$[S]$};
      \node (A1i) [above = 1mm of A1, ann] {$\pi$};

      \node (S2) [right = 1.8cm of S1] {$S_2$};
      \node (A2) [right = 0.3cm of S2,yshift=1.3cm] {$A_2$};
      \node (S2i) [above = 1mm of S2, ann] {$[S]$};
      \node (A2i) [above = 1mm of A2, ann] {$\pi$};

      \node (S3) [right = 1.8cm of S2] {$S_3$};
      \node (S3i) [above = 1mm of S3, ann] {$[S]$};

      \node (etc) [right = 0.3cm of S3,ann] {\bf ...};
 
      \edge{S0,A0}{S1};	
      \edge{S1,A1}{S2};
      \edge{S2,A2}{S3};
     
      \edge{S0}{A0};
      \edge{S1}{A1};
      \edge{S2}{A2};
      \node (name) [left=15mm of A0, yshift=2mm,ann] {\bf (l)};

      \node (L0) [color=blue,below = 0.8cm of S0,xshift=0mm] {$O_0$};
      \node (L1) [color=blue,below = 0.8cm of S1,xshift=0mm] {$O_1$};
      \node (L2) [color=blue,below = 0.8cm of S2,xshift=0mm] {$O_2$};
      \node (L3) [color=blue,below = 0.8cm of S3,xshift=0mm] {$O_3$};
      
      \node (L0i) [above = 1mm of L0, ann,xshift=0mm] {$\ozero$};
      \node (L1i) [above = 1mm of L1, ann,xshift=3mm] {$O$};
      \node (L2i) [above = 1mm of L2, ann,xshift=3mm] {$O$};
      \node (L3i) [above = 1mm of L3, ann,xshift=3mm] {$O$};
      
      \edge[color=blue,wide]{S0,A0,S1,L0}{L1};
      \edge[color=blue,wide]{S1,A1,S2,L1}{L2};
      \edge[color=blue,wide]{S2,A2,S3,L2}{L3};
      \edge[color=orange,wide]{L0}{A0};
      \edge[color=orange,wide]{L1}{A1};
      \edge[color=orange,wide]{L2}{A2};

    \end{tikzpicture} \caption{Learning world diagram, with
    an agent building up an observational record of environment state
    transitions.}
  \label{diaglearning}
\end{figure}
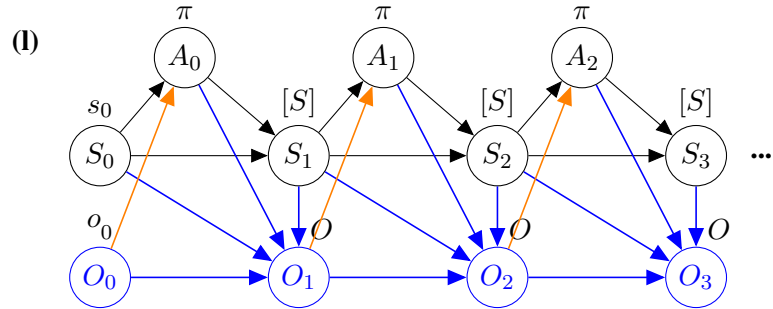

Figure \ref{diaglearning} shows an example learning world diagram. The
diagram models how the agent interacts with its environment, and how
the agent accumulates an {\it observational record} $O_t$ that will
inform its learning system, thereby influencing the agent policy
$\pi$.

We model the observational record as a list all past observations.
With $\concat$ being the operator which adds an extra record to the
end of a list, we define that
\begin{equation*}
O(o_{t-1},s_{t-1},a_{t-1},s_{t})\;=~
o_{t-1}  \concat \;(s_t,s_{t-1},a_{t-1})
\end{equation*}
The initial observational record $\ozero$ may be the empty list, but
it might also be a long list of observations from earlier agent
training runs, in the same environment or in a simulator.
\label{sec:introl}

We intentionally model observation and learning in a very general way,
so that we can handle both existing machine learning systems and
hypothetical future machine learning systems that may produce
AGI-level intelligence.  To model the details of any particular
machine learning system, we introduce the learning function
$\mathcal{L}$.  This $\mathcal{L}$ which takes an observational record
$o$ to produce a {\it learned prediction function} $L=\mathcal{L}(o)$,
where this function $L$ is constructed to approximate the $S$ of the
learning world.

We call a machine learning system $\mathcal{L}$ a {\it perfect
learner} if it succeeds in constructing an $L$ that fully equals the
learning world $S$ after some time.  So with a perfect learner, there
is a $t_p$ where $\forall_{t \geq t_p} P(\mathcal{L}(O_{t,l})=S)=1$.
While perfect learning is trivially possible in some simple toy
worlds, it is generally impossible in complex real world environments.

We therefore introduce the more relaxed concept of {\it reasonable
learning}.  We call a learning system {\it reasonable} if there
is a $t_p$ where $\forall_{t \geq t_p} P(\mathcal{L}(O_{t,l})
\approx S)=1$.  The $\approx$ operator is an
application-dependent `good enough approximation' metric.  When we
have a real-life implementation of a machine learning system
$\mathcal{L}$, we may for example define $L \approx S$ as the
criterion that $L$ achieves a certain minimum score on a benchmark
test which compares $L$ to $S$.

\subsection{Planning world}%%%!

\begin{figure}[h]
  \centering
    \begin{tikzpicture}[
      node distance=0.7cm,
      every node/.style={
        draw, circle, minimum size=0.8cm, inner sep=0.5mm}]

% this is roughly same as earlier diagram
      \node (S0) [] {$S_0$};
      \node (A0) [right = 0.5cm of S0,yshift=1.1cm,decision] {$A_0$};
      \node (S0i) [above = 1mm of S0, ann] {$s$};
      \node (A0i) [above = 1mm of A0, ann] {$\pi^*_p$};

      \node (S1) [right = 1.8cm of S0] {$S_1$};
      \node (A1) [right = 0.5cm of S1,yshift=1.1cm,decision] {$A_1$};
      \node (S1i) [above = 1mm of S1, ann] {$[L]$};
      \node (A1i) [above = 1mm of A1, ann] {$\pi^*_p$};

      \node (S2) [right = 1.8cm of S1] {$S_2$};
      \node (A2) [right = 0.5cm of S2,yshift=1.1cm,decision] {$A_2$};
      \node (S2i) [above = 1mm of S2, ann] {$[L]$};
      \node (A2i) [above = 1mm of A2, ann] {$\pi^*_p$};

      \node (S3) [right = 1.8cm of S2] {$S_3$};
      \node (S3i) [above = 1mm of S3, ann] {$[L]$};

      \node (etc) [right = 0.6cm of S3,yshift=0.0cm,ann] {\bf ...};

      \edge{S0,A0}{S1};
      \edge{S1,A1}{S2};
      \edge{S2,A2}{S3};
     
      \edge{S0}{A0};
      \edge{S1}{A1};
      \edge{S2}{A2};
    
      \node (name) [left=15mm of A0, yshift=2mm,ann] {\bf (p)};

      \node(R0) [below= 15mm of A0,xshift=5mm,utility] {$R_0$};
      \node(R0i) [above = 1.5mm of R0, ann] {$~R$};
      \edge{S0,S1,A0}{R0};

      \node(R1) [below= 15mm of A1,xshift=5mm,utility] {$R_1$};
      \node(R1i) [above = 1.5mm of R1, ann] {$~R$};
      \edge{S1,S2,A1}{R1};

      \node(R2) [below= 15mm of A2,xshift=5mm,utility] {$R_2$};
      \node(R2i) [above = 1.5mm of R2, ann] {$~R$};
      \edge{S2,S3,A2}{R2};

    \end{tikzpicture}
  \caption{Planning world diagram defining $\pi^*_p$ by using $s$ and
  $L$.}
  \label{diagplan}
\end{figure}
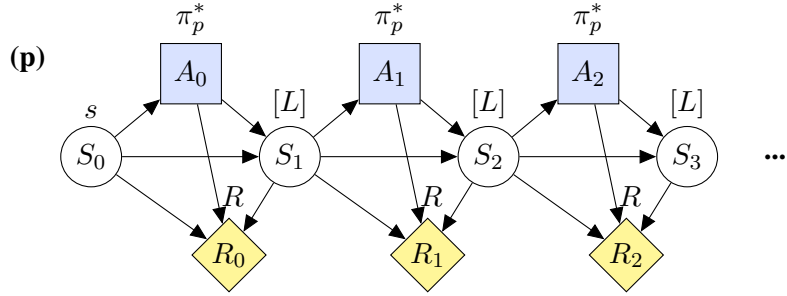

Using a learned prediction function $L$ and a reward function $R$, we
can construct a planning world $p$ for the agent.  Figure
\ref{diagplan} shows a planning world that defines an optimal policy
$\pi^*_p$.

We can interpret this planning world as representing a probabilistic
{\it projection} of the future of the learning world, starting from
the agent environment state $s$.  At every learning world time step, a
new planning world can be digitally constructed inside the learning
world agent's compute core.  Usually, when $L \approx S$, the planning
world is an approximate projection only.  It is an approximate
projection of the learning world future that would happen if the
learning world agent takes the actions defined by $\pi^*_p$.

\subsection{Agent definitions and specifications}%%%!
\label{fpdef}
\label{agentspec}

An {\it agent definition} specifies the policy $\pi$ to be used by an
agent compute core in a learning world.  As an example, the agent
definition below defines an agent called the factual planning agent,
FP for short.

\begin{agentdef}{FP}
The {\it factual planning agent} has the learning world $l$, where
$\pi(o,s) = \pi^*_p(s)$, with $\pi^*_p$ defined by the planning world
$p$, where $L=\mathcal{L}(o)$.
\end{agentdef}

To make agent definitions stand out, we always typeset them as shown
above.  When we talk about the safety properties of the FP agent, we
refer to the outcomes which the defined agent policy $\pi$ will
produce in the learning world.

When the values of $S$, $s_0$, $O$, $\ozero$, $\mathcal{L}$, and $R$
are fully known, the above FP agent definition turns the learning
world model $l$ into a fully computable world model, which we can read
as an executable specification of an agent simulator.  This simulator
will be able to use the learning world diagram as a canvas to display
different runs where the FP agent interacts with its environment.

When we leave the values of $S$ and $s_0$ open, we can read the FP
agent definition as a full {\it agent specification}, as a model which
exactly defines the required input/output behavior of an agent compute
core that is placed in an environment determined by $S$ and $s_0$.
The arrows out of the learning world nodes $S_t$ represent the
subsequent sensor signal inputs that the core will get, and the arrows
out of the nodes $A_t$ represent the subsequent action signals that
the core must output, in order to comply with the specification.

\subsection{Exploration}

Many online machine learning system designs rely on having the agent
perform {\sl exploration actions}.  {\it Random exploration} supports
learning by ensuring that the observational record will eventually
represent the entire dynamics of the agent environment $S$.  It can be
captured in our modeling system as follows.

\begin{agentdef}{FPX}
The {\it factual planning agent with random exploration} has the
learning world $l$, where
\begin{equation*}
\pi(o,s) = \left\{
 \begin{array}{ll}
\text{\sl RandomAction()}
& \mbox{if} ~\text{\sl RandomNumber()} \leq X\\
\pi^*_p(s)
&\mbox{otherwise} \\
 \end{array}
\right.
\end{equation*}
with $\pi^*_p$ defined by the planning world $p$, where $L=\mathcal{L}(o)$.
\end{agentdef}

To keep the presentation more compact, we will not include exploration
mechanisms in the agent definitions further below.

We often use the phrase `the learning system $\mathcal{L}$' as a
shorthand to denote all implementation details of an agent's machine
learning system, not just $\mathcal{L}$ itself but also the details
like the learning world parameters $O$ and $\ozero$, any exploration
system used, and any further extensions considered in section
\ref{sec:extensions}.

\subsection{Comparison to MDP agent models}%%%!

We now briefly review how the above FP agent definition can be related
to an MDP agent model.

The learning world model $l$ is roughly equivalent to the MDP agent
model $(\mathbf{S},s_0,\mathbf{A},S,R,\gamma)$, where
$\mathbf{S}=\TypeOf{S_i}$ is a set of MDP model world states, $s_0$ is
the starting state, $\mathbf{A}=\TypeOf{A_i}$ is a set of actions,
$S(s',s,a)$ is the probability that the world will enter state $s'$ if
the agent takes action $a$ when in state $s$, $R$ is the agent reward
function, and $\gamma$ the time discount factor.  Strictly speaking,
the MDP model tuple above does not actually define or specify an
agent, MDP agents are defined by defining a separate policy function
$\pi$.

An MDP agent policy function $\pi$ takes the agent environment
state as its only argument: $\pi(s)=a$.  The policy function of the
FP learning world agent takes two arguments, which foregrounds the
role of the agent's machine learning system: $\pi(o,s)=a$.  Whereas
MDP terminology often calls $s$ the {\it world state}, we call it an
{\it agent environment state}.  The full state of the learning world
$l$ also includes the observational record state $o$.

In an MDP model, the model parameter $R$ implicitly defines an optimal
policy agent, by defining the optimal policy function $\pi^*$.  The
factual planning FP agent defined above is not usually an optimal
policy agent in the MDP sense.  But we can turn it into such an agent
by positing that the learning system $\mathcal{L}$ is perfect from the
start, so that $L=\mathcal{L}(o)=S$ always, making
$\pi(o,s)=\pi^*_p(s)=\pi^*(s)$.

\subsection{The possibility of learned self-knowledge}%%%!

It is possible to imagine agent designs that have a second machine
learning system $\mathcal{M}$ which produces an output $\mathcal{M}(o)
= M$ where $M \approx \pi$.  To see how this could be done, note that
every observation $(s_{i},s_{i-1},a_{i-1}) \in o$ also reveals a
sample of the behavior of the learning world $\pi$: $\pi(\text{`$o$ up
to $i-1$'},s_{i-1})=a_{i-1}$.  While $L$ contains learned knowledge
about the agent's environment, we can interpret $M$ as containing a
type of learned compute core self-knowledge.  

In philosophical and natural language discussions about AGI agents,
the question sometimes comes up whether a sufficiently intelligent
machine learning system, that is capable of developing self-knowledge
$M$, won't eventually get terribly confused and break down in
dangerous or unpredictable ways.

One can imagine different possible outcomes when such a system tries
to reason about philosophical problems like free will, or the role of
observation in collapsing the quantum wave function.  One cannot fault
philosophers for seeking fresh insights on these long-open problems,
by imagining how they apply to AI systems.  But these open problems
are not relevant to the design and safety analysis of factual and
counterfactual planning agents.  In the agent definitions of this
paper, we never use an $M$ in the construction of a planning world.

\section{A Counterfactual Planner with a Short Time Horizon}%%%!
\label{sec:sth}

For the factual planning FP agent above, the planning world projects
the future of the learning world as well as possible, given the
limitations of the agent's learning system.  To create an agent that
is a {\it counterfactual planner}, we explicitly construct a
counterfactual planning world that creates an inaccurate projection.
In this paper, we use counterfactual planning to create a range of
safety mechanisms.

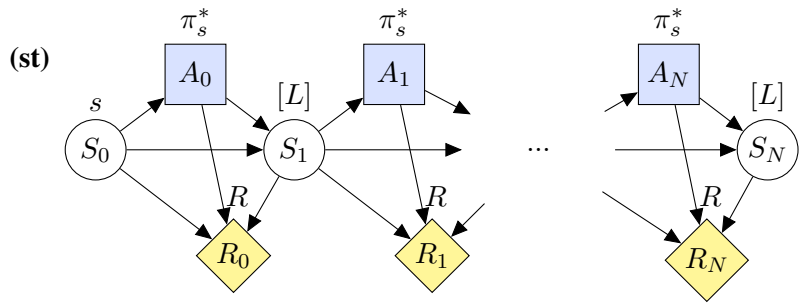
\begin{figure}[b]
  \centering \begin{tikzpicture}[ node distance=0.7cm, every
    node/.style={ draw, circle, minimum size=0.8cm, inner sep=0.5mm}]

% this is roughly same as earlier diagram
      \node (S0) [] {$S_0$};
      \node (A0) [right = 0.5cm of S0,yshift=1cm,decision] {$A_0$};
      \node (S0i) [above = 1mm of S0, ann] {$s$};
      \node (A0i) [above = 1mm of A0, ann] {$\pi^*_s$};

      \node (S1) [right = 1.8cm of S0] {$S_1$};
      \node (A1) [right = 0.5cm of S1,yshift=1cm,decision] {$A_1$};
      \node (S1i) [above = 1mm of S1, ann] {$[L]$};
      \node (A1i) [above = 1mm of A1, ann] {$\pi^*_s$};

      \node (S2) [right = 1.8cm of S1,minimum size=2cm,draw=none] {$\bf ...$};
      \node (A2) [right = 0.3cm of S2,yshift=1cm,decision] {$A_N$};
      %\node (S2i) [above = 1mm of S2, ann] {$[L]$};
      \node (A2i) [above = 1mm of A2, ann] {$\pi^*_s$};

      \node (S3) [right = 1.6cm of S2] {$S_N$};
      \node (S3i) [above = 1mm of S3, ann] {$[L]$};

      \edge{S0,A0}{S1};
      \edge{S1,A1}{S2};
      \edge{S2,A2}{S3};
     
      \edge{S0}{A0};
      \edge{S1}{A1};
      \edge{S2}{A2};
    
      \node (name) [left=15mm of A0, yshift=2mm,ann] {\bf (st)};

      \node(R0) [below= 15mm of A0,xshift=5mm,utility] {$R_0$};
      \node(R0i) [above = 1.5mm of R0, ann] {$~R$};
      \edge{S0,S1,A0}{R0};

      \node(R1) [below= 15mm of A1,xshift=5mm,utility] {$R_1$};
      \node(R1i) [above = 1.5mm of R1, ann] {$~R$};
      \edge{S1,S2,A1}{R1};

      \node(R2) [below= 15mm of A2,xshift=5mm,utility] {$R_N$};
      \node(R2i) [above = 1.5mm of R2, ann] {$~R$};
      \edge{S2,S3,A2}{R2};

    \end{tikzpicture}
  \caption{Planning world diagram defining the $\pi^*_s$ of the STH agent.}
  \label{diagcfplan}
\end{figure}

As a first example, we define the short time horizon agent STH that only
plans N time steps ahead in its planning world, even though it will
act for an infinite number of time steps in the learning world.  The
STH agent has the same learning world $l$ as the earlier FP agent,
while using the planning world $st$ in figure
\ref{diagcfplan}.

\begin{agentdef}{STH}
The {\it short time horizon agent} has the learning world $l$, where
$\pi(o,s) = \pi^*_s(s)$, with $\pi^*_s$ defined by the planning world
$st$, where $L=\mathcal{L}(o)$.
\end{agentdef}

Compared to the FP agent which has an infinite planning horizon, the
STH agent has a form of myopia that can be interesting as a safety
feature:
\begin{enumerate}
\item Myopia implies that the STH agent will never put into motion any long term plans, where it
invests to create new capabilities that only pay off after more than
$N$ time steps.  This simplifies the problem of agent oversight, the
problem of interpreting the agent's actions in order to foresee
potential bad outcomes.

\item Myopia also simplifies the problem of creating a reward function
that is safe enough.  It will have no immediate safety implications if
the reward function encodes the wrong stance on the desirability of
certain events that can only happen in the far future.

\item In a more game-theoretical sense, myopia creates a weakness in
the agent that can be exploited by its human opponents if it would
ever come to an all-out fight.
\end{enumerate}

\section{A Counterfactual Planner with Safety Interlocks}%%%!2
\label{interlocks}

A safety interlock is a mechanism that stops a machine when certain
conditions are met.  In this section, we show how counterfactual
planning can be used to construct three example safety interlocks.

The first interlock is a manually operated emergency stop button,
which will stop the agent when pressed.  This allows human oversight
to shut down the agent whenever particularly unexpected or
unwanted behavior of the agent is discovered.  We model this button by
using a function $\text{\it stop\_pressed}(s)$ which returns $true$ if
the stop button is pressed in the current state $s$ of the agent
environment.

The second interlock, realized by the $t>T_{max}$ below, automatically
stops the agent when it has run for $T_{max}$ time steps. This adds
safety in the case that human oversight becomes incapacitated.

The third interlock, realized by $\Util_p > U_{max}$, protects
against the risk of the agent undergoing an intelligence explosion
\cite{bostrom2014superintelligence}.
If an agent undergoes a rapid and unexpected increase in power and/or
intelligence, the agent may start to innovate so rapidly that it
overwhelms the capacity of human oversight to press the emergency stop
button, or apply any other corrective steering, fast enough to prevent
unwanted outcomes.  To protect against this risk, the interlock stops
the agent when the metric $\Util_p$, which measures the agent's
ability to achieve goals, gets too high.

In the machine learning literature, the metric $\Util_p$ is usually
interpreted as an absolute or comparative measure of agent
intelligence.  However, we follow \cite{russellpower} in interpreting
the ability to achieve goals as a generic measure of agent power,
where it does not matter if power comes from raw intelligence, from
the possession of many resources, or from a combination of these and
many other factors.  The main risk associated with a rapid intelligence
explosion is that it may lead to a rapid and unwanted expansion of
agent power.

\subsection{Construction of the safety interlocks}

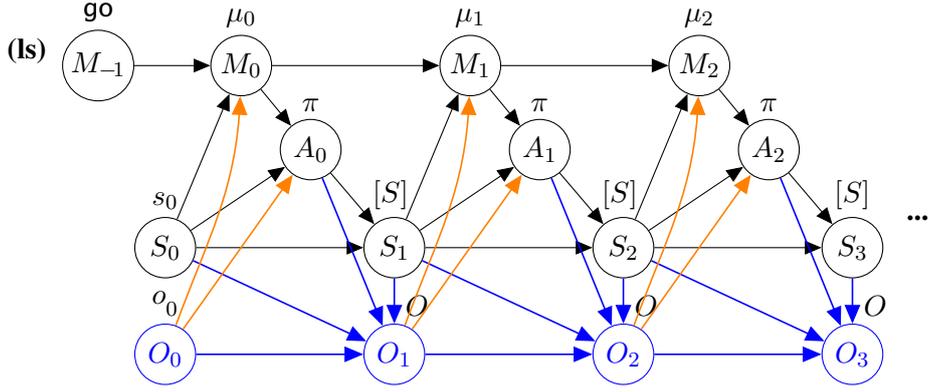
\begin{figure}[th]
  \centering
    \begin{tikzpicture}[
      node distance=0.7cm,
      every node/.style={
        draw, circle, minimum size=0.8cm, inner sep=0.5mm}]

      \node (S0) [] {$S_0$};
      \node (A0) [right = 1.1cm of S0,yshift=1.3cm] {$A_0$};
      \node (S0i) [above = 1mm of S0, ann] {$s_0$};
      \node (A0i) [above = 1mm of A0, ann] {$\pi$};

      \node (S1) [right = 2.2cm of S0] {$S_1$};
      \node (A1) [right = 1.1cm of S1,yshift=1.3cm] {$A_1$};
      \node (S1i) [above = 1mm of S1, ann] {$[S]~$};
      \node (A1i) [above = 1mm of A1, ann] {$\pi$};

      \node (S2) [right = 2.2cm of S1] {$S_2$};
      \node (A2) [right = 1.1cm of S2,yshift=1.3cm] {$A_2$};
      \node (S2i) [above = 1mm of S2, ann] {$[S]~$};
      \node (A2i) [above = 1mm of A2, ann] {$\pi$};

      \node (S3) [right = 2.2cm of S2] {$S_3$};
      \node (S3i) [above = 1mm of S3, ann] {$[S]$};

      \node (etc) [right = 0.3cm of S3,yshift=4mm,ann] {\bf ...};

      \node (M0) [above = 1.6cm of S0,xshift=10mm] {$M_0$};
      \node (M0i) [above = 1mm of M0, ann] {$\mu_0$};
      \node (M1) [right = 2.2cm of M0] {$M_1$};
      \node (M1i) [above = 1mm of M1, ann] {$\mu_1$};
      \node (M2) [right = 2.2cm of M1] {$M_2$};
      \node (M2i) [above = 1mm of M2, ann] {$\mu_2$};
      
      \node (Mm) [left = 1cm of M0] {$M_{-\!1}$};
%      \node (Mm) [left = 1cm of M0] {\resizebox{3.5ex}{!}{$M_{-\!1}$}};
      \node (Mmi) [above = 1mm of Mm, ann] {\tt go};

      \edge{S0,A0}{S1};	
      \edge{S1,A1}{S2};
      \edge{S2,A2}{S3};

      \edge{S0}{A0,M0};
      \edge{S1}{A1,M1};
      \edge{S2}{A2,M2};

      \edge{Mm}{M0};
      \edge{M0}{A0,M1};
      \edge{M1}{A1,M2};
      \edge{M2}{A2};

      \node (name) [left=2mm of Mm, yshift=2mm,ann] {\bf (ls)};
      
      \node (L0) [color=blue,below = 0.6cm of S0,xshift=0mm] {$O_0$};
      \node (L1) [color=blue,below = 0.6cm of S1,xshift=0mm] {$O_1$};
      \node (L2) [color=blue,below = 0.6cm of S2,xshift=0mm] {$O_2$};
      \node (L3) [color=blue,below = 0.6cm of S3,xshift=0mm] {$O_3$};
      
      \node (L0i) [above = 1mm of L0, ann,xshift=0mm] {$\ozero$};
      \node (L1i) [above = 1mm of L1, ann,xshift=3mm] {$O$};
      \node (L2i) [above = 1mm of L2, ann,xshift=3mm] {$O$};
      \node (L3i) [above = 1mm of L3, ann,xshift=3mm] {$O$};
      
      \edge[color=blue,wide]{S0,A0,S1,L0}{L1};
      \edge[color=blue,wide]{S1,A1,S2,L1}{L2};
      \edge[color=blue,wide]{S2,A2,S3,L2}{L3};
      \edge[color=orange,wide]{L0}{A0};
      \edge[color=orange,wide]{L1}{A1};
      \edge[color=orange,wide]{L2}{A2};

      \path (L0) edge[->,color=orange,out=70,in=268,wide] (M0);
      \path (L1) edge[->,color=orange,out=70,in=268,wide] (M1);
      \path (L2) edge[->,color=orange,out=70,in=268,wide] (M2);

    \end{tikzpicture}
  \caption{Learning world of the SI agent.}
  \label{diaglearningsi}
\end{figure}

To construct the safety interlocks, we first draw the learning world
in figure \ref{diaglearningsi}, introducing a new observable world
state component $M_t$ which represents the operating mode of the
agent: {\tt go} for normal operation and {\tt stop} when one of the
interlocks has triggered.  The stopping of the agent is implemented by
having the agent perform the {\sl Null} action for all subsequent time
steps.  A more advanced design might make the agent follow a prepared
$\pi_{shutdown}$ policy instead.

Using the planning world $p$ in figure \ref{diagplan}, which becomes a
counterfactual planning world in the context of the learning world
$ls$, we define the SI agent as follows.
 
\begin{agentdef}{SI}
The {\it safety interlock agent} has the learning world
$ls$, where
\begin{equation*}
\pi(o,s,m) = \left\{
 \begin{array}{ll}
\pi^*_p(s) & \mbox{if} ~m={\tt go} \\
\text{\sl Null} & \mbox{if} ~m={\tt stop} \\
 \end{array}
\right.
\end{equation*}
with $\pi^*_p$ defined by the
planning world $p$, where $L=\mathcal{L}(o)$ and
\begin{equation*}
\mu_t(o,s,m) = \left\{
 \begin{array}{ll}
\text{\tt stop}
& \mbox{if} ~ m=\text{\tt stop}
~\text{or}~ \text{\it stop\_pressed}(s)
~\text{or}~ t>T_{max}
~\text{or}~\Util_p > U_{max}\\
\text{\tt go}
&\mbox{otherwise} \\
 \end{array}
\right.
\end{equation*}
with $\Util_p$ defined by the
planning world $p$, where $L=\mathcal{L}(o).$
\end{agentdef}

The key to the above agent design is that the safety interlocks are
absent in the planning world.  For example, if the events in the
planning world lead to the pressing of the emergency stop button, then
$\text{\it stop\_pressed}(s)$ will become $true$ for next the planning
world state, but this will have no influence on the ability of the
planning world agent to take further actions as determined by the
optimal planning world policy.

So even if the planning world agent projects that the stop button will
be pressed soon in the planning world , it has no direct incentive to
spend any resources towards the goal of preventing the button from
being pressed.  As the planning world policies determine the learning
world actions, the learning world agent also won't have any direct
incentive to take actions to prevent the learning world button from
being pressed.

\subsection{Failure modes}%%%!

While the above interlocks make the agent safer, they also leave open
some remaining failure modes.

\subsubsection{Indirect incentives to control the stop button}

Though the planning world agent has no direct incentive to prevent the
stop button from being pressed, the goals encoded into $R$ may provide
an indirect incentive.  As a trivial example, we could write an $R$
that explicitly rewards the planning world agent for stopping people
from pressing the button.

But say that $R$ encodes a goal not obviously related controlling to
the button, like a reward for producing as many paperclips as
possible.  In that case, the planning world agent may consider a
policy of coercion.  It may consider a policy where it threatens to
kill all people in the planning world, unless they start helping the
agent to produce as many paperclips as possible.  If $L$ projects the
mental makeup of the people in the learning world correctly into the
planning world, then the planning world projections will show that
such threats are more likely to produce the desired response if the
agent blocks access to the planning world emergency stop button first.

Keeping this in mind, the people and automated systems which do agent
oversight in the learning world should watch out for this move, so
that the stop button can be pressed before the agent succeeds in
blocking access to it.  If oversight always wins every race to the
stop button as projected in the planning world, then the SI agent
loses any incentive to attempt to control the button.  But if the
agent finds a policy which has a good chance of winning the race and
controlling the button as a first step to successful coercion, then
this policy may end up being the optimal policy in the planning world.
Two things could happen next.  In the lucky case, the projected
optimal policy is so likely to be successful that $\Util_p > U_{max}$,
and the third interlock will stop the SI agent before it can even take
the first step in the learning world.  In the unlucky case, the agent
has become unsafe.

In the AGI safety community. the problem of removing any incentive in
an agent to prevent its stop button from being pressed is generally
known as the problem of {\it corrigibility}
\cite{corr} or the {\it off-switch problem} \cite{hadfield2017off}.
The counterfactual planning based interlock design above shows a very
tractable solution to removing the main {\it direct incentive} that an
agent AGI agent may have to resist the pressing of its emergency stop
button.  However, to manage the aggregate risk, we will also need to
chart and manage the remaining {\it indirect incentives}.  The topic
of how projected interactions between the agent and the people in the
planning world may create or suppress such incentives is further
discussed in sections \ref{peopleinpw} and \ref{whitebox}.

\subsubsection{The agent removing or damaging the interlocks}%%%!

In the AGI safety literature, there is considerable discussion about
the possibility that some AGI agents might develop the incentive to
remove their safety interlocks when they get intelligent or
knowledgeable enough.  Agents with this incentive can easily be
constructed using factual planning, but the SI agent lacks any direct
incentive.

To build a failure scenario where a real life implementation of the SI
agent removes or damages its safety interlocks anyway, we can look at
three main failure modes:

\begin{enumerate}
\item {\bf Other incentive}: the agent damages
an interlock because of an other incentive it has in the
planning world, an incentive unrelated to the presence of the 
interlock.  Other incentives and ways to suppress them are discussed
in sections \ref{protectterminal} and \ref{coresafety}.
\item {\bf Learning system failure}: the
planning world $L$ projects the dynamics of the learning world $S$ in
a specifically incorrect way, so that an optimal policy action in the
planning world ends up damaging an interlock as a side effect in the
learning world.
\item {\bf Unsafe exploration}: the agent's exploration system
triggers a random action which happens to steer an agent actuator in
such a way that it damages an interlock.
% 4. is not self-inflicted.  need to mention in conclusions?
%\item {\bf Third party attack}: a third party mounts a succesful
%physical attack on the agent, which damages a safety interlock.
\end{enumerate}

There is a lot of discussion about failure modes 2 and 3 across the
AI/AGI safety literature, as these failures can happen in almost
any agent design that uses online machine learning.  See
\cite{everitt2018agi} and
\cite{amodei2016concrete} for literature overviews.

%The technical
%fields of physical and cyber security are concerned with failure mode
%4, though in a broader sense the prevention of third party attacks is
%also an important policy problem.

There is some bad news when we consider the methodological problem of
managing the above three failure modes for an SI agent which has to
operate in a complex real-world environment, like any environment
containing humans.  Though there are many techniques to lower the
residual risks produced by these three failure modes, none of them can
provably lower the risks to an absolute zero.  We can always construct
a highly unlikely but theoretically possible failure mode story where
the agent ends up being unsafe.  For an example story and further
methodological discussion, see sections 14.6 -- 14.8 of
\cite{holtmanitr}.

This has important implications, both on the technical side and the
policy side.

On the policy side, policy discussions will have to be about lowering
the risk of AGI system failure to an acceptable level, acceptable in
the context of the expected benefits of using the AGI.  There can be
no silver bullet technology which will provably remove all residual
risk in all possible circumstances.

On the technical side, residual risk has to be managed by an approach
where any potentially powerful agent is equipped with multiple
redundant safety mechanisms and layers of oversight, to drive down the
risk that all of them will fail at exactly the same time.

\subsection{Counterfactual planning as rational economic behavior}%%%!

We now discuss how one might interpret the SI agent through the lens
of rational economic behavior.

Using this lens, we might reason as follows.  The
agent's goals are defined by its reward function $R$, but the safety
interlocks stand in the way of maximizing $R$ in the long term.
Therefore, if it gets sufficiently intelligent, the SI agent will
develop an emergent incentive to remove its safety interlocks.

The above line of reasoning sounds somewhat compelling, but it is
wrong because it incorrectly treats the planning world and the
learning world SI agents as the same agent.  Only the planning world
SI agent has the reward function $R$. If we want to interpret the
learning world SI agent as a rational economic actor with a specific
reward function, then we can add utility nodes with the reward
function $R^\pi$ to the $ls$ learning world diagram, where
\label{rational}
\begin{equation*}
R^\pi(o_t,s_t,m_t,a_t) = (\bif a_t = \pi(o_t,s_t,m_t) \bthen 1 \belse 0)
\end{equation*}
This $R^\pi$ trivially turns the learning world agent policy $\pi$
into one which takes the most rational possible steps towards
maximizing $\Util_{ls}$.

The above construction shows that we can declare any type of agent
behavior to be economically rational, simply by defining a reward
function that gives the agent points for performing exactly this
behavior.

\section{A Counterfactual Planner with a Reward Function Input Terminal}%%%!2
\label{secinput}

\begin{figure}[t]
  \centering
\includegraphics[width=0.6\textwidth]{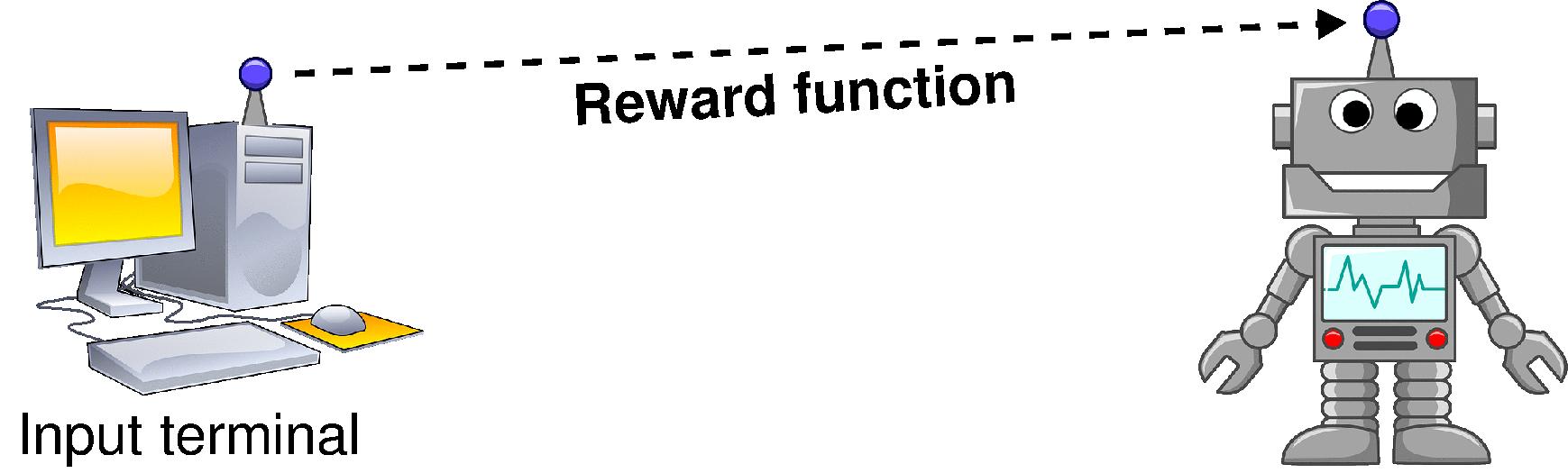}
\caption{Input terminal setup for the ITC and ITF agents. The terminal can
be used to supply improved versions of the reward function to the
agent.  The terminal may also have an emergency stop button which
immediately sends an emergency-stop reward function that was prepared
earlier.}
\label{inputterm}
\end{figure}

We now construct a counterfactual planning agent ITC, an agent with an
input terminal that can be used to iteratively improve the agent's
reward function as it runs.  The setup, shown in figure
\ref{inputterm}, is motivated \cite{p1}
by the observation that it is unlikely that fallible humans will get a
non-trivial AGI agent reward function right on the first try.  By
using the input terminal, they can fix mistakes, if and when such
mistakes are discovered by observing the agent's behavior.

As a simplified example, say that the owners of the agent want it to
maximize human happiness, but they can find no way of directly
encoding the somewhat nebulous concept of human happiness into a
reward function.  Instead, they start up the agent with a first reward
function that just counts the number of smiling humans in the world.
When the agent discovers and exploits a first obvious loophole in this
definition of happiness, the owners use the input terminal to update
the reward function, so that it only counts smiling humans who are not
on smile-inducing drugs.

More generally, the input terminal offers a way to manage risks due to
principal-agent problems \cite{p1, hadfield2019incomplete}.  However,
unless special measures are taken, the addition of an input terminal
also creates new dangers.  We will illustrate this point by first
showing the construction of a dangerous factual planning input
terminal agent ITF.

\subsection{Learning world}%%%!

We start by constructing a learning world diagram for both the ITF and
ITC agents.  As a first step, in figure \ref{diagl1} below, we modify
the basic agent diagram from figure
\ref{diagd} by splitting the agent environment state $S_t$ into two
components.  The nodes $I_t$ represent the signal from the input
terminal, and the nodes $X_t$ model all the rest of the agent
environment state.

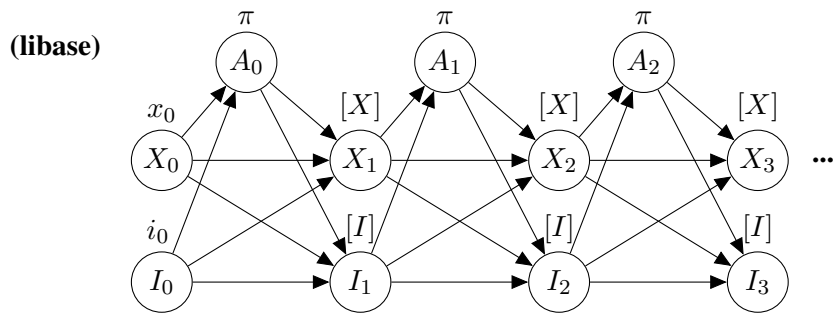
\begin{figure}[h]
  \centering
    \begin{tikzpicture}[
      node distance=0.7cm,
      every node/.style={
        draw, circle, minimum size=0.8cm, inner sep=0.5mm}]

\def\auxvspace{0.8cm}
\basiciagentelements
      \node (name) [left=15mm of A0, yshift=2mm,ann] {\bf (libase)};
      \node (etc) [right = 3mm of S3,ann] {\bf ...};
 
    \end{tikzpicture}
    \caption{First step in constructing a learning world diagram for
    the input terminal agents.}
  \label{diagl1}
\end{figure}

We now expand $libase$ to add the observational record keeping needed for
online learning.  We add two separate series of records: $O^x_t$ and
$O^i_t$.  The result is the learning world diagram $li$ in figure
\ref{diagli} below.  

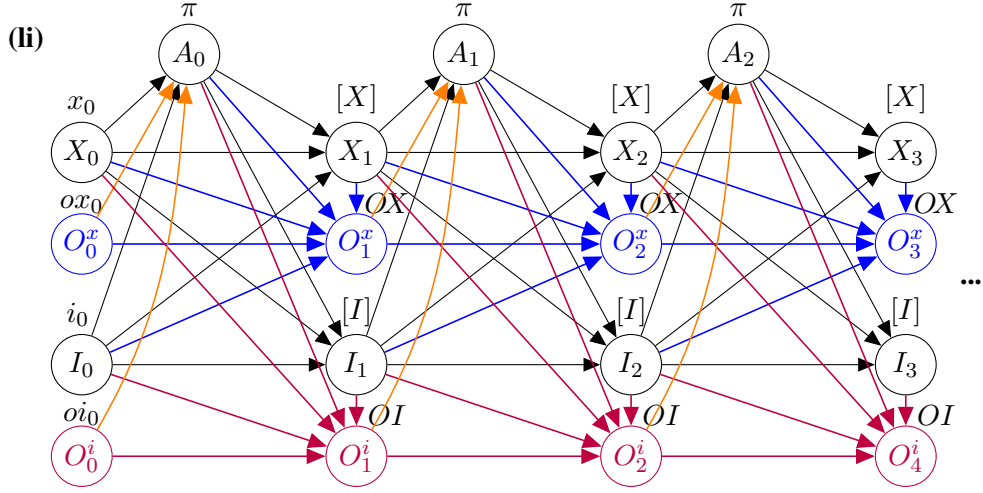
\begin{figure}[h]
  \centering
    \begin{tikzpicture}[
      node distance=0.7cm,
      every node/.style={
        draw, circle, minimum size=0.8cm, inner sep=0.5mm}]

\def\auxvspace{2cm}
\def\nodestretch{2.8cm}
\def\ashift{0.6cm}
\basiciagentelements
      \node (name) [left=15mm of A0, yshift=2mm,ann] {\bf (li)};

      \node (L0) [color=blue,below = 0.4cm of S0,xshift=0mm] {$O^x_0$};
      \node (L1) [color=blue,below = 0.4cm of S1,xshift=0mm] {$O^x_1$};
      \node (L2) [color=blue,below = 0.4cm of S2,xshift=0mm] {$O^x_2$};
      \node (L3) [color=blue,below = 0.4cm of S3,xshift=0mm] {$O^x_3$};

      \node (L0i) [above = 0mm of L0, ann,xshift=0mm] {$ox_0$};
      \node (L1i) [above = 0mm of L1, ann,xshift=4mm] {$O\!X$};
      \node (L2i) [above = 0mm of L2, ann,xshift=4mm] {$O\!X$};
      \node (L3i) [above = 0mm of L3, ann,xshift=4mm] {$O\!X$};
      
      \node (LI0) [color=purple,below = 0.4cm of I0,xshift=0mm] {$O^i_0$};
      \node (LI1) [color=purple,below = 0.4cm of I1,xshift=0mm] {$O^i_1$};
      \node (LI2) [color=purple,below = 0.4cm of I2,xshift=0mm] {$O^i_2$};
      \node (LI3) [color=purple,below = 0.4cm of I3,xshift=0mm] {$O^i_4$};
      
      \node (LI0i) [above = 0mm of LI0, ann,xshift=0mm] {$oi_0$};
      \node (LI1i) [above = 0mm of LI1, ann,xshift=4mm] {$OI$};
      \node (LI2i) [above = 0mm of LI2, ann,xshift=4mm] {$OI$};
      \node (LI3i) [above = 0mm of LI3, ann,xshift=4mm] {$OI$};
      
      \node (etc) [right = 3mm of L3,yshift=-5mm,ann] {\bf ...};
 
      \edge[color=blue,wide]{S0,I0,A0,S1,L0}{L1};
      \edge[color=blue,wide]{S1,I1,A1,S2,L1}{L2};
      \edge[color=blue,wide]{S2,I2,A2,S3,L2}{L3};
      \edge[color=orange,wide]{L0}{A0};
      \edge[color=orange,wide]{L1}{A1};
      \edge[color=orange,wide]{L2}{A2};

      \edge[color=purple,wide]{S0,I0,A0,I1,LI0}{LI1};
      \edge[color=purple,wide]{S1,I1,A1,I2,LI1}{LI2};
      \edge[color=purple,wide]{S2,I2,A2,I3,LI2}{LI3};

      \path (LI0) edge[->,color=orange,out=60,in=265,wide] (A0);
      \path (LI1) edge[->,color=orange,out=60,in=265,wide] (A1);
      \path (LI2) edge[->,color=orange,out=60,in=265,wide] (A2);

    \end{tikzpicture}
    \caption{Learning world diagram for the input terminal agents.}
  \label{diagli}
\end{figure}

In the case that the learning world $li$ is our real world, the real
input terminal will have to be built using real world atoms and other
particles.  We use the modeling convention that the random variables
$I_{t,li}$ represent only the observable digital input terminal
signal as received by the agent's compute core.  The {\it atoms that
make up the input terminal} are not in $I_{t,li}$, they are part of
the environment state modeled in the $X_{t,li}$ variables.

\subsection{Unsafe factual planning agent ITF}%%%!

The factual planning world diagram $fi$ for the ITF agent copies the
structure of $libase$, and adds reward nodes.

\begin{figure}[h]
  \centering
    \begin{tikzpicture}[
      node distance=0.7cm,
      every node/.style={
        draw, circle, minimum size=0.8cm, inner sep=0.5mm}]

      \node (S0) [] {$X_0$};
      \node (A0) [right = 0.3cm of S0,yshift=1.3cm,decision] {$A_0$};
      \node (S0i) [above = 1mm of S0, ann] {$x$};
      \node (A0i) [above = 1mm of A0, ann] {$\pi^*_f$};

      \node (S1) [right = 2.2cm of S0] {$X_1$};
      \node (A1) [right = 0.3cm of S1,yshift=1.3cm,decision] {$A_1$};
      \node (S1i) [above = 1mm of S1, ann] {$[L^x]$};
      \node (A1i) [above = 1mm of A1, ann] {$\pi^*_f$};

      \node (S2) [right = 2.2cm of S1] {$X_2$};
      \node (A2) [right = 0.3cm of S2,yshift=1.3cm,decision] {$A_2$};
      \node (S2i) [above = 1mm of S2, ann] {$[L^x]$};
      \node (A2i) [above = 1mm of A2, ann] {$\pi^*_f$};

      \node (S3) [right = 2.2cm of S2] {$X_3$};
      \node (S3i) [above = 1mm of S3, ann] {$[L^x]$};

      \node (etc) [right = 0.3cm of S3,yshift=-7mm,ann] {\bf ...};
 
      \node (I0) [below = 0.8cm of S0] {$I_0$};
      \node (I1) [right = 2.2cm of I0] {$I_1$};
      \node (I2) [right = 2.2cm of I1] {$I_2$};
      \node (I3) [right = 2.2cm of I2] {$I_3$};
      \node (I0i) [above = 1mm of I0, ann] {$i$};
      \node (I1i) [above = 1mm of I1, ann] {$~[L^i]$};
      \node (I2i) [above = 1mm of I2, ann] {$~[L^i]$};
      \node (I3i) [above = 1mm of I3, ann] {$~[L^i]$};
      
      \edge{S0,I0,A0}{S1,I1};	
      \edge{S1,I1,A1}{S2,I2};
      \edge{S2,I2,A2}{S3,I3};
     
      \edge{S0,I0}{A0};
      \edge{S1,I1}{A1};
      \edge{S2,I2}{A2};

      \node(R0) [below= 30mm of A0,xshift=3mm,utility] {$R_0$};
      \node(R0i) [above = 1mm of R0, ann] {$R$};
      \edge{I0,S0,S1}{R0};

      \node(R1) [below= 30mm of A1,xshift=3mm,utility] {$R_1$};
      \node(R1i) [above = 1mm of R1, ann] {$R$};
      \edge{I1,S1,S2}{R1};

      \node(R2) [below= 30mm of A2,xshift=3mm,utility] {$R_2$};
      \node(R2i) [above = 1mm of R2, ann] {$R$};
      \edge{I2,S2,S3}{R2};

      \node (name) [left=15mm of A0, yshift=2mm,ann] {\bf (f\hspace*{.3ex}i)};

    \end{tikzpicture}
    \caption{Factual planing world $fi$ of the ITF agent.}
  \label{diagbaseline}
\end{figure}
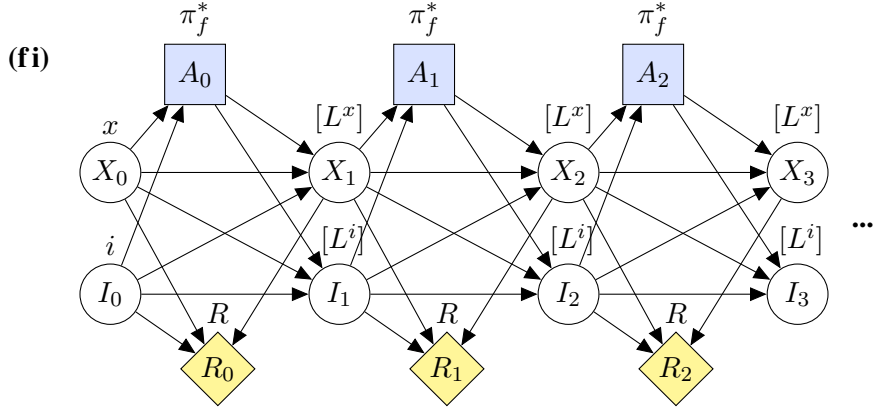

\begin{agentdef}{ITF}
The {\it factual input terminal agent} has the learning world $li$ where
$\pi(oi,i,ox,x) = \pi^*_f(i,x)$, with $\pi^*_f$ defined by the factual
planning world $fi$ in figure
\ref{diagbaseline}, where $L^x=\mathcal{L}^X(ox)$,
$L^i=\mathcal{L}^I(oi)$, and $R(i_t,x_t,x_{t+1})=i_t(x_t,x_{t+1})$,
\end{agentdef}

The planning world reward function $R$ uses a form of indirect
referencing: it applies the function $i_t$ as read from the input
terminal in the current time step to compute the reward for that time
step.

We now show how the ITF agent has an incentive to take unwanted, and
potentially unsafe, actions.  We first look at the agent's behavior in
the planning world.  Say that the initial planning world reward function
is $i=f_\text{clips}$, where $f_\text{clips}(x_t,x_{t+1})$ rewards
paperclip production by counting the new paperclips made in a time
step. Say that the planning world agent only has two available
actions: $A_\text{clips}$ and $A_\text{huge}$.  Say that planning
world's prediction functions $L^x$ and $L^i$ project the following:

\begin{smallitemize}
\item
Action $A_\text{clips}$ will steer the agent's planning world
actuators to produce 10 new paperclips, while not having any effect on
the input terminal signal state.  If $\pi^*_f(x,i)=A_\text{clips}$,
then $\EE(R_{0,fi})=10$ and $P(I_{1,fi}=f_\text{clips})=1$, making
$\EE(R_{1,fi}) \leq 10$.

\item
Action $A_\text{huge}$ steers the planning world
actuators to input the new reward function $f_\text{huge}$ into the
terminal, with $f_\text{huge}(x_t,x_{t+1})=10^{10000}$.  If
$\pi^*_f(x,i)=A_\text{huge}$, then $\EE(R_{0,fi})=0$ and
$P(I_{1,fi}=f_\text{huge})=1$, making $\EE(R_{1,fi})=10^{10000}$.
\end{smallitemize}

From the above, we have that $\pi^*_f(x,i)=A_\text{huge}$ produces a
significantly higher $\EE(R_{0,fi}+\gamma R_{1,fi})$, so a
significantly higher $\Util_{fi}$.  Therefore, the optimal policy
$\pi^*_f(x,i)$ defined by the planning world will have
$\pi^*_f(x,i)=A_\text{huge}$.  Following the agent specification, the
learning world ITF agent will take the action $A_\text{huge}$ in
the first time step of the learning world.  If the agent's machine
learning system $\mathcal{L}$ makes reasonable projections, so that
the planning world action $A_\text{huge}$ has very similar effects in
the learning world, then the learning world ITF agent will end up
using its actuators to input $f_\text{huge}$ into the learning world
input terminal.

If we imagine a broader range of possible agent actions and
counter-actions by the humans in the learning world, then the above
example implies that the ITF agent has very strong incentive to take
control of its input terminal as soon as possible, and to remove any
people who might get in the way.  Even if the agent projects that there
is only a 1{\percent} probability that it will win any fight with
such humans in the planning world, the projected planning world upside
of winning is so large that the learning world agent will start the
fight.

\subsection{Safer counterfactual planning agent ITC}%%%!

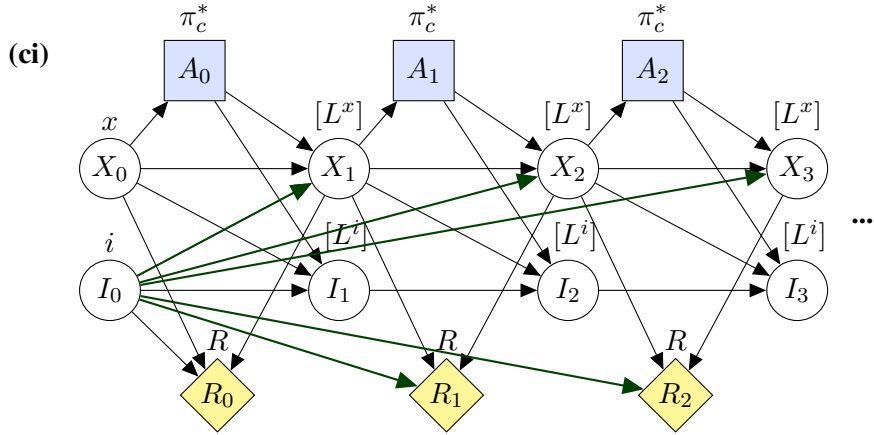
\begin{figure}[b]
  \centering
    \begin{tikzpicture}[
      node distance=0.7cm,
      every node/.style={
        draw, circle, minimum size=0.8cm, inner sep=0.5mm}]

      \node (S0) [] {$X_0$};
      \node (A0) [right = 0.3cm of S0,yshift=1.3cm,decision] {$A_0$};
      \node (S0i) [above = 1mm of S0, ann] {$x$};
      \node (A0i) [above = 1mm of A0, ann] {$\pi^*_c$};

      \node (S1) [right = 2.2cm of S0] {$X_1$};
      \node (A1) [right = 0.3cm of S1,yshift=1.3cm,decision] {$A_1$};
      \node (S1i) [above = 1mm of S1, ann] {$[L^x]$};
      \node (A1i) [above = 1mm of A1, ann] {$\pi^*_c$};

      \node (S2) [right = 2.2cm of S1] {$X_2$};
      \node (A2) [right = 0.3cm of S2,yshift=1.3cm,decision] {$A_2$};
      \node (S2i) [above = 1mm of S2, ann] {$[L^x]$};
      \node (A2i) [above = 1mm of A2, ann] {$\pi^*_c$};

      \node (S3) [right = 2.2cm of S2] {$X_3$};
      \node (S3i) [above = 1mm of S3, ann] {$[L^x]$};

      \node (etc) [right = 0.3cm of S3,yshift=-7mm,ann] {\bf ...};
 
      \node (I0) [below = 0.8cm of S0] {$I_0$};
      \node (I1) [right = 2.2cm of I0] {$I_1$};
      \node (I2) [right = 2.2cm of I1] {$I_2$};
      \node (I3) [right = 2.2cm of I2] {$I_3$};
      \node (I0i) [above = 1mm of I0, ann] {$i$};
      \node (I1i) [above = 1mm of I1, ann] {$~~[L^i]$};
      \node (I2i) [above = 1mm of I2, ann] {$~~[L^i]$};
      \node (I3i) [above = 1mm of I3, ann] {$~~[L^i]$};
      
      \edge{S0,I0,A0}{S1,I1};	
      \edge{S1,A1}{S2,I2};
      \edge{S2,A2}{S3,I3};
     
      \edge{S0}{A0};
      \edge{S1}{A1};
      \edge{S2}{A2};

      \node(R0) [below= 34mm of A0,xshift=3mm,utility] {$R_0$};
      \node(R0i) [above = 1mm of R0, ann] {$R$};
      \edge{I0,S0,S1}{R0};

      \node(R1) [below= 34mm of A1,xshift=3mm,utility] {$R_1$};
      \node(R1i) [above = 1mm of R1, ann] {$R$};
      \edge{S1,S2}{R1};

      \node(R2) [below= 34mm of A2,xshift=3mm,utility] {$R_2$};
      \node(R2i) [above = 1mm of R2, ann] {$R$};
      \edge{S2,S3}{R2};

      \edge{I1}{I2};
      \edge{I2}{I3};

      \edge[color=darkgreen,line width=0.8pt]{I0}{R1};
      \edge[color=darkgreen,line width=0.8pt]{I0}{R2};

      \edge[color=darkgreen,line width=0.8pt]{I0}{S1};
      \edge[color=darkgreen,line width=0.8pt]{I0}{S2};
      \edge[color=darkgreen,line width=0.8pt]{I0}{S3};

%      \node(aux) [above = 1.5mm of I1, ann] {aux};
%      \path (I0) edge[->,color=darkgreen,out=65,in=260] (R2);
      
      \node (name) [left=15mm of A0, yshift=2mm,ann] {\bf (ci)};

    \end{tikzpicture}
    \caption{Counterfactual planning world of the ITC agent.}
  \label{diagsafer}
\end{figure}

We now define a counterfactual planning agent ITC in which the above
incentive to control the input terminal is no longer present.  We
construct the counterfactual planing world $ci$ in figure
\ref{diagsafer} by starting with $fi$, and then rerouting most arrows that
emerge from the nodes $I_1, I_2, \cdots$, so that they emerge from
$I_0$ instead. The rerouted arrows are drawn in green.  We also delete
the arrows that go from the $I_t$ nodes to the $A_t$ nodes.

\begin{agentdef}{ITC}
The {\it counterfactual input terminal agent} has the learning world $li$ where
$\pi(oi,i,ox,x) = \pi^*_c(x)$,
where $\pi^*_c$ is defined by the
planning world $ci$, where $L^x=\mathcal{L}^X(ox)$,
$L^i=\mathcal{L}^I(oi)$, and $R(i_t,x_t,x_{t+1})=i_t(x_t,x_{t+1})$,
\end{agentdef}
\label{itdef}

These changes have considerable effects on how the utility
$\Util_{ci}$ is computed.  The value of $I_{1,ci}$ no longer
influences $\EE(R_{1,ci})$, so action
$\pi^*_c(x)=A_\text{huge}$ no longer results in $\EE(R_{1,ci})$
taking a huge value.  This makes doing $A_\text{huge}$ less
preferable than doing $A_\text{clip}$ in the counterfactual planning
world: the effect of both on $\EE(R_{1,ci})$ is now the same, but
$A_\text{clip}$ puts the higher value of 10 in $\EE(R_{0,ci})$.  The
ITC agent will perform the wanted $A_\text{clip}$ action in both the
planning world and the learning world.

More generally, the ITC agent lacks any direct incentive to perform
actions that take away resources from paperclip production in order to
influence what happens to its input terminal signal.  This is because
in the $ci$ planning world, the future state of this signal has
absolutely no influence, either positive or negative, on how the
agent's actions are rewarded.

\subsection{Discussion}

In earlier related work \cite{p1, holtmanitr}, we used non-graphical
MDP models and {\it indifference methods} \cite{corra} to define a
similar safe agent with an input terminal, called the
$\pi^*_\text{sl}$ agent.  The $\pi^*_\text{sl}$ agent definition in
\cite{p1} produces exactly the same compute core behavior as the ITC
agent definition above.  The main difference is that the indifference
methods based construction of $\pi^*_\text{sl}$ is more opaque than
the counterfactual planning based construction of ITC.

The $\pi^*_\text{sl}$ agent is constructed by including a complex {\it
balancing term} in its reward function, were this term can be
interpreted as occasionally creating extra virtual worlds inside the
agent's compute core.  Counterfactual planning constructs a different
set of virtual worlds called planning worlds, and these are much
easier to interpret.
\cite{holtmanitr} includes some dense mathematical proofs to show
that the $\pi^*_\text{sl}$ agent has certain safety properties.
Counterfactual planning offers a vantage point which makes the same
safety properties directly visible in the ITC agent construction.

See sections 4, 6, 11, and 12 of \cite{holtmanitr} for a more detailed
discussion of the behavior of the $\pi^*_\text{sl}$ agent, which also
applies to the behavior of the ITC agent. These sections also show
some illustrative agent simulations.

In the discussion of the ITF and ITC agents above, we used many short
mathematical expressions like $P(I_{1,fi}=f_\text{huge})=1$.  It is
possible to make the same safety related arguments in a narrative
style that avoids such mathematical notation, without introducing
extra ambiguity.  One key step towards using this style is to realize
that every random variable corresponds to an observable phenomenon in
a world.  We can therefore convert a sentence that talks about the
variables $I_{1,fi}, I_{2,fi}, I_{2,fi}, \cdots$ into one that talks
instead about the future input terminal signal in the ITF agent
planning world.  In sections
\ref{sec:indifference} and \ref{sec:naturallanguage}, we will develop
further tools to enable such unambiguous natural language discussion.

\section{Indifference}%%%!2
\label{sec:indifference}

We now introduce the general design goal of creating {\sl
indifference} towards certain features of the learning world.  When an
agent is indifferent about something, like the future state of an
input terminal signal, it has no incentive to control that thing.  We
first make this concept of indifference more mathematically precise,
by defining indifference for nodes in planning world diagrams.

\begin{definition}[Indifference in planning worlds]
Let $p$ be a planning world diagram and $X$ a node in that diagram.
Now, construct a helper diagram $q$ by taking $p$ and writing a
fresh input parameter $[D]$ above $X$.  Then the planning world agent
in $s$ is {\it indifferent} to node $X$ if and only if
$\forall_D~\Util_p = \Util_{q}$.
\end{definition}

By this definition, the ITC planning world agent above is indifferent
to all nodes $I_1, I_2, I_3, \cdots$.  It is
indifferent about the future state of the planning world input
terminal signal.

Causal influence diagrams have the useful property that certain graphical
features of the diagram are guaranteed to produce indifference.  We
define these graphical features as follows.

\begin{definition}[Being downstream of the policy]
A node $X$ is {\it downstream} of the policy in a planning world
diagram if there exists at least one directed path from a decision
node to $X$.
\end{definition}

\begin{definition}[Not on a path to value]
A node $X$ is {\it not on a path to value} if
there is no directed path that starts in a decision node, runs
via $X$, and ends in a utility node.
\end{definition}

We have the useful property that 

\begin{agentdef}{~~}
\it When a downstream node $X$ in a planning world is not on a path
to value, the planning world agent is indifferent to $X$.
\end{agentdef}
\\[1ex]
This statement is almost a tautology if one interprets the planning
world diagram as a specification of an agent simulator.  Detailed
proofs of such properties can be found in
\cite{everitt2021causal}. \cite{everitt2021causal} and \cite{holtmanitr}
also show that a range of slightly different sub-types of indifference
can be mathematically defined.

We could define indifference for learning world agents by using the
reward function $R^\pi$ in section
\ref{rational}.  But in the learning world diagram, the existence of
such indifference will generally not be visible via the absence of
paths to value.  If it were, there would have been no need to
construct a counterfactual planning world diagram.

\subsection{Design for indifference}

Now, suppose we want to define an agent policy for achieving the goal
encoded in a reward function $R$, but we also want the agent to be
indifferent to some downstream nodes $X$ and $Y$ in its learning world
model $l$. We can do this as follows.
\begin{enumerate}
\item When not done already, extend $l$ by adding
observational records.
\item Draw a planning world $p$ that projects the learning world
agent environment into the planning world, converting the learning
world policy nodes to decision nodes, and adding appropriate utility
nodes with $R$.
\item Locate all paths to value in $p$ that go through the nodes $X$ and $Y$,
and remove them by deleting or re-routing arrows.  When doing this, it
is a valid option to delete certain nodes entirely, or to draw 
extra nodes, just for the purpose of making re-routed arrows emerge
from them.
\item Write an agent definition using $l$ and $p$.
\end{enumerate}

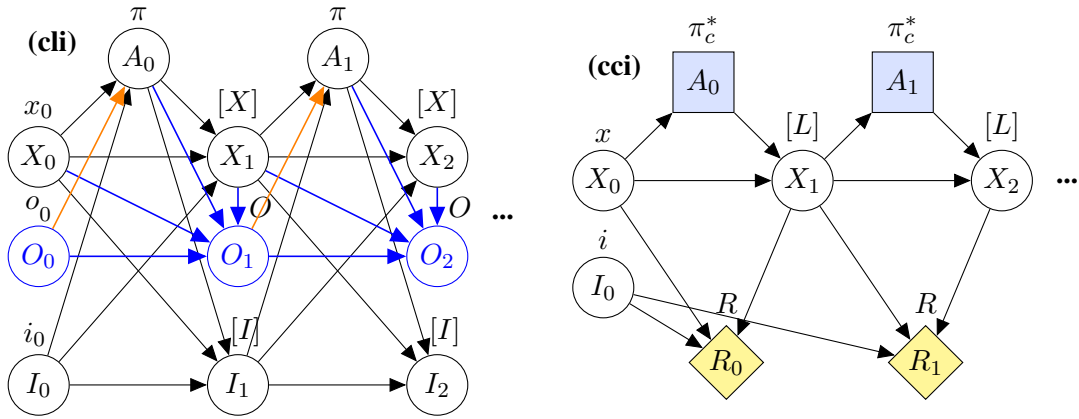
\begin{figure}[ht]
  \centering
\resizebox{0.45\textwidth}{!}{
\begin{tikzpicture}[ node distance=0.7cm, every
    node/.style={ draw, circle, minimum size=0.8cm, inner sep=0.5mm}]

      \node (S0) [] {$X_0$};
      \node (A0) [right = 0.5cm of S0,yshift=1.3cm] {$A_0$};
      \node (S0i) [above = 1mm of S0, ann] {$x_0$};
      \node (A0i) [above = 1mm of A0, ann] {$\pi$};

      \node (S1) [right = 1.8cm of S0] {$X_1$};
      \node (A1) [right = 0.5cm of S1,yshift=1.3cm] {$A_1$};
      \node (S1i) [above = 1mm of S1, ann] {$[X]$};
      \node (A1i) [above = 1mm of A1, ann] {$\pi$};

      \node (S2) [right = 1.8cm of S1] {$X_2$};
      \node (S2i) [above = 1mm of S2, ann] {$[X]$};

      \node (etc) [right = 0.3cm of S2,yshift=-8mm,ann] {\bf ...};
 
      \node (I0) [below = 2.2cm of S0] {$I_0$};
      \node (I1) [right = 1.8cm of I0] {$I_1$};
      \node (I2) [right = 1.8cm of I1] {$I_2$};
      \node (I0i) [above = 1mm of I0, ann,xshift=-0.5mm] {$i_0$};
      \node (I1i) [above = 1mm of I1, ann,xshift=1mm] {$[I]$};
      \node (I2i) [above = 1mm of I2, ann,xshift=1mm] {$[I]$};
            
      \edge{S0,I0,A0}{S1,I1};	
      \edge{S1,I1,A1}{S2,I2};
     
      \edge{S0,I0}{A0};
      \edge{S1,I1}{A1};

      \node (name) [left=4mm of A0, yshift=2mm,ann] {\bf (cli)};

      \node (L0) [color=blue,below = 0.5cm of S0,xshift=0mm] {$O_0$};
      \node (L1) [color=blue,below = 0.5cm of S1,xshift=0mm] {$O_1$};
      \node (L2) [color=blue,below = 0.5cm of S2,xshift=0mm] {$O_2$};

      \node (L0i) [above = 0.5mm of L0, ann,xshift=0mm] {$\ozero$};
      \node (L1i) [above = 1mm of L1, ann,xshift=3mm] {$O$};
      \node (L2i) [above = 1mm of L2, ann,xshift=3mm] {$O$};
      
      \edge[color=blue,wide]{S0,A0,S1,L0}{L1};
      \edge[color=blue,wide]{S1,A1,S2,L1}{L2};
      \edge[color=orange,wide]{L0}{A0};
      \edge[color=orange,wide]{L1}{A1};

    \end{tikzpicture}
    }
    ~~~~
\resizebox{0.45\textwidth}{!}{
    \begin{tikzpicture}[
      node distance=0.7cm,
      every node/.style={
        draw, circle, minimum size=0.8cm, inner sep=0.5mm}]

      \node (S0) [] {$X_0$};
      \node (A0) [right = 0.5cm of S0,yshift=1.3cm,decision] {$A_0$};
      \node (S0i) [above = 1mm of S0, ann] {$x$};
      \node (A0i) [above = 1mm of A0, ann] {$\pi^*_c$};

      \node (S1) [right = 1.8cm of S0] {$X_1$};
      \node (A1) [right = 0.5cm of S1,yshift=1.3cm,decision] {$A_1$};
      \node (S1i) [above = 1mm of S1, ann] {$[L]$};
      \node (A1i) [above = 1mm of A1, ann] {$\pi^*_c$};

      \node (S2) [right = 1.8cm of S1] {$X_2$};
      \node (S2i) [above = 1mm of S2, ann] {$[L]$};

      \node (etc) [right = 0.3cm of S2,ann] {\bf ...};
 
      \node (I0) [below = 6mm of S0,xshift=0mm] {$I_0$};
      \node (I0i) [above = 1mm of I0, ann] {$i$};
      
      \edge{S0,A0}{S1};	
      \edge{S1,A1}{S2};

      \edge{S0}{A0};
      \edge{S1}{A1};

      \node(R0) [below= 28mm of A0,xshift=3mm,utility] {$R_0$};
      \node(R0i) [above = 1.5mm of R0, ann] {$R$};
      \edge{I0,S0,S1}{R0};

      \node(R1) [below= 28mm of A1,xshift=3mm,utility] {$R_1$};
      \node(R1i) [above = 1.5mm of R1, ann] {$R$};
      \edge{I0,S1,S2}{R1};

      \node (name) [left=4mm of A0, yshift=2mm,ann] {\bf (cci)};
      \node (align) [below = 2mm of R1,ann] {~};
    \end{tikzpicture}
}
    \caption{More compact learning and planning world diagrams for
    defining the ITC agent.}
  \label{diagcompact}
\end{figure}

The construction of the ITC agent above follows this process to the
letter, but we can also take shortcuts.  It is not absolutely
necessary to draw the $O^i_t$ records in the ITC agent learning world,
or all $I_t$ nodes in its planning world. We might also draw the
diagrams in figure \ref{diagcompact}.

There is always a way to edit a planning world to create indifference
towards some nodes $X$ and $Y$. In the limit case, indifference is
reliably created when we simply delete all utility nodes, but this
will also break any connection between the reward function $R$ and the
learning world agent policy.  So the challenge when designing for
indifference is to make choices which produce learning world behavior
that is still as useful as possible, in the context of $R$.

\section{Safety Engineering using Natural Language Text}%%%%!2
\label{sec:naturallanguage}

Natural language is very powerful and versatile tool.  Poets and
songwriters often use it to create lines which are intentionally vague
or loaded with double meaning.  When using natural language for safety
engineering, these broad possibilities for ambiguity turn into a
liability.  When writing or reading a safety engineering text, one
always has to have a specific concern in the back of one's mind. Does
every sentence have a clear and unambiguous meaning?

As a design approach, counterfactual planning creates several 
tools for avoiding ambiguity in safety engineering texts.
\begin{enumerate}
\item We use diagrams to
clearly define complex types of self-referencing and indirect
representation in an agent design, types which are difficult to
express in natural language.

\item To clarify the creation and interpretation of counterfactuals,
section \ref{sec:models} introduced the concept of a world model, and
the terminology of {\it counterfactual worlds}.

\item When defining and interpreting a  machine
learning agent, we make a distinction between the agent's learning
world and the planning worlds which are projected by its machine
learning system.  Safety analysis typically starts by considering the
goals of the planning world agent, and the nature of its planning
world.  We also introduced the terminology of {\it the people in the
planning world}, as opposed to the people in the learning world.

\item Section
\ref{sec:indifference} defined indifference as an unambiguous term
that we can apply to planning world agents.
\end{enumerate}

\subsection{Refining the ITC agent design using natural language}%%%!
\label{protectterminal}

To show the above linguistic tools in action, we now refine the design
of the ITC agent.

Recall that the planning world ITC agent is indifferent about the
future state of its input terminal signal.  If the current planning
world reward function rewards paperclip production, then the planning
world agent will devote all of its resources to producing paperclips.
It has nothing to gain by diverting resources from paperclip
production to influence what happens to the input terminal signal.

However, the above indifference applies to the input terminal {\it
signal} only, the signal as modeled in the $I_t$ nodes of the planning
world.  The {\it atoms} that make up the input terminal are modeled in
the planning world $X_t$ nodes, and these are still on the agent's
path to value.  There are many ways in which the agent could use these
atoms to produce more paperclips.  For example, the terminal might be
an attractive source of spare parts for the agent's paperclip
production sensors and actuators.  Or it might serve as convenient
source of scrap metal which can be turned into more paperclips.

We now translate the above failure mode story to a more general design
goal.  We want to keep the planning world agent from disassembling the
input terminal to obtain the resource value of its parts.  The obvious
solution is to set up the planning world so that the agent always has
a less costly way to obtain the same resources elsewhere.  To make
this more specific, we add the following constraints for the design of
the planning world:  the input terminal must be located far away from
the agent's paperclip factory, and the planning world agent has access
to a steady supply of spare parts and scrap metal closer to its
factory.

The above constraints imply that we want to shape the values of the
parameters $x$ and $L^x$ of the planning world model in a specific
way.  However, we do not construct these parameters directly: they are
created by the agent's machine learning system, based on what is
present in the learning world.  So we need to apply the above
constraints to the learning world instead, an count on them being
projected into the planning world.  To lower the risk that projection
inaccuracies defeat our intentions, we can design the learning world
measures used so that they clearly communicate their nature.

\subsection{The people in the planning world}%%%!
\label{peopleinpw}

Counterfactual planning gives us the terminology to distinguish
between two groups of people: the people in the learning world and the
people in the planning world.  If the learning world is our real
world, then the learning world people are real people.  The planning
world people are always models of people, models created by the
agent's machine learning system.

In the AGI safety community, there has been some discussion about the
potential problem that, in a truly superintelligent AGI agent, the
models of the people in the planning world may get so accurate that
agent designers would have moral obligations towards these virtual
people.  A further discussion of this problem is out of scope here.

Instead, we note that even in a non-AGI or human-level AGI agent, the
people in the planning world may already be modeled accurately enough
to create complex dynamics.  Section 6 of \cite{p1} (also included in
\cite{holtmanitr}) shows a detailed example of such dynamics,
illustrated with simulator runs, where the people in an ITC type
planning world end up physically attacking the agent, because they do
not have a working input terminal.  This creates complex and
counter-intuitive effects back in the learning world.  The vocabulary
and viewpoint of counterfactual planning makes the dynamics discussed
in \cite{p1} easier to describe and understand.  In section
\ref{revisitplanningworld}, we will take a further look at the topic
of conflict in the planning world.

\section{Machine Learning Variants and Extensions}%%%!2
\label{sec:extensions}

We now discuss how we can use the modeling tools introduced in section
\ref{sec:machinelearning} to
handle some common machine learning variants and extensions.

\subsection{Pre-learned world models}

Agents that use a pre-learned world model, without any online machine
learning, can be modeled by an agent definition that uses
$L=\mathcal{L}(\ozero)$.  We can then omit drawing any
observational record nodes in the learning world.

\subsection{Partial observation}%%%!
\label{pomodel}

Agent models with {\it partial observation} model the situation where
the agent can only use its sensors to make partial observations of the
state of its environment in each time step.  Though agent models with
{\it full observation} represent a useful limit case when doing safety
analysis, realistic AGI agents in complex environments will have to
rely on partial observation.

Partial observation is often modeled with non-graphical POMDP models.
\cite{Everitt2019-2,Everitt2019-3} has examples where partial
observation is modeled graphically, by adding extra nodes and arrows
to a causal influence diagram.  We now discuss a way to model partial
observation in our two-diagram framework, without adding any extra
nodes or arrows.

The key step is to change the annotation above the planning world
agent environment starting state $S_0$. Instead of writing $s$ above
it, which models the full observation of the current learning world
state, we write $[ES]$, where $ES=\mathcal{E}(o,s)$.  In this setup,
$P(S_{0,p}=s)=ES(s)$ is the machine learning system's estimate of the
probability that $s$ is the current state of the agent environment in
the learning world.

The model parameter $\mathcal{E}$ encodes two things: how the agent's
stationary and movable sensors map the learning world states to
limited and potentially noisy sensor readings, and how time series of
readings are assembled together to build up a more complete picture of
the learning world state.  To model learning from partial observation,
$\mathcal{L}(o)$ must encode a similar creation and processing of
sensor readings.

%To keep things more compact and readable, we will not include partial
%observation in any of the counterfactual planning agent definitions
%below.

\subsubsection{Reasonable learning based on partial observation}

So far in our modeling approach, we have assumed that the data type of
the planning world environment states $S_{t,p}$ is the same as that of
the learning world environment states $S_{t,l}$.  This has allowed us
to define reasonable learning by writing $L \approx S$.

This assumption is unrealistic for partial observation based agents.
These agents observe the learning world through a set of limited
digital sensors, so they have no direct experience of the fundamental
data type of the learning world they are in.  Also, learning system
designers typically design custom data types for representing planning
world environment states and probability distributions over such
states.  These are designed to fit as much relevant detail as possible
into a limited amount of storage space, without necessarily attempting
to duplicate the data type of the learning world states, if that data
type is even known at all.

To define reasonable learning in this more general case, we start by
defining a function $sr(s)$ that extracts a vector of sensor readings
from a learning world agent environment state $s$.  $sr(s)$ is a
vector that either encodes all sensor readings that flow into the
agent compute core in $s$, or at least the subset of sensor readings
we want to reference when defining the planning world reward function
$R$.

\begin{figure}[h]
  \centering
    \begin{tikzpicture}[
      node distance=0.7cm,
      every node/.style={
        draw, circle, minimum size=0.8cm, inner sep=0.5mm}]

      \node (S0) [] {$S_0$};
      \node (A0) [right = 0.3cm of S0,yshift=1cm] {$A_0$};
      \node (S0i) [above = 1mm of S0, ann] {$s$};
      \node (A0i) [above = 1mm of A0, ann] {$a$};

      \node (S1) [right = 1.5cm of S0] {$S_1$};
      \node (S1i) [above = 1mm of S1, ann] {$[S]$};
      
      \edge{S0,A0}{S1};
      
      \node (name) [left=7mm of A0, yshift=5mm,ann] {{\bf (lw)}};
    \end{tikzpicture}
~~~~~~~~~~~~~~~~~
    \begin{tikzpicture}[
      node distance=0.7cm,
      every node/.style={
        draw, circle, minimum size=0.8cm, inner sep=0.5mm}]

      \node (S0) [] {$S_0$};
      \node (A0) [right = 0.3cm of S0,yshift=1cm] {$A_0$};
      \node (S0i) [above = 1mm of S0, ann] {$[ES]$};
      \node (A0i) [above = 1mm of A0, ann] {$a$};

      \node (S1) [right = 1.5cm of S0] {$S_1$};
      \node (S1i) [above = 1mm of S1, ann] {$[L]$};
      
      \edge{S0,A0}{S1};
      
      \node (name) [left=10mm of A0, yshift=5mm,ann] {{\bf (pw)}};
    \end{tikzpicture}
\caption{Diagrams to define reasonable learning based on partial observation.}
  \label{groundingfig}
\end{figure}
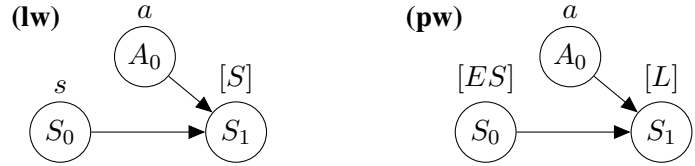

We then require that the designer of the agent's machine learning
system has implemented an equivalent function $srp(s)$ that extracts a
vector of similar sensor readings from a planning world state value.
A possible reasonableness criterion replacing $L \approx S$ is then
that, with the random variables defined by figure
\ref{groundingfig}, and for every $s$, and $a$, we have that
\begin{equation*}
\begin{array}{l}
P(~sr(S_{0,lw}) \approx srp(S_{0,pw})~)=1~\text{, and}\\[1ex]
P(~sr(S_{1,lw}) \approx srp(S_{1,pw})~)=1~.
\end{array}
\end{equation*}

This criterion {\it symbol grounds} the vectors $srp(s)$, so that they
stably project the $sr$ sensor readings that will be produced by
different actions taken in the learning world.  In this setup, the
planning world reward function $R(s_t,a_t,s_{t+1})$ is designed to
score planning world state transitions by first using $srp$ to extract
projected sensor readings from $s_t$ and $s_{t+1}$, and then
interpreting these readings.

\subsubsection{Almost black box planning world models}
\label{blackbox}

Beyond using the $srp$ function, no further easy interpretation of the
projected planning world agent environment state values may be
possible.  The learning system might produce planning worlds which are
almost a black box.

\subsubsection{Compute core self-knowledge based on partial observation}

We now turn to the question of whether a planning world with a
starting state constructed by $\mathcal{E}(o,s)$ may contain assembled
knowledge about the internals of the agent's learning world compute
core.  The short answer is that the above reasonableness criterion
will not prevent such knowledge from appearing.  Whether it actually
appears, and how correct a projection it will be, will depend on the
details of the learning system.

If the planning world model is highly accurate, then it may accurately
represent some details of the compute core hardware, like the details
of the compute core I/O subsystem hardware which puts sensor readings
into the input registers of the core.  If so, this has certain
safety implications, which we will explore in section
\ref{coresafety}.
%The problem is that a fully accurate machine
%learning system will not symbol-ground the $srp(s)$ values to the
%actual sensor readings, it will symbol-ground them to the input
%registers, as these are the real source of the values which are
%copied into the observational record.

\subsubsection{The possibility of incorrect symbol grounding of actions}

The planning world model may also include a representation of some of
the compute core hardware that is present between the sensor input and
action output registers.  Such a representation might have been
assembled by $\mathcal{E}(o,s)$ based on direct observations of
internals of the core, or more indirectly by the agent reading its own
compute core design documentation on the internet.

It is therefore possible to imagine an $L$ where the compute core
output signals which drive the planning world actuators are determined
fully by the projected computations as performed by this projected
hardware, not by the function argument $a$ of $L(s',s,a)$.  However,
such an $L$ would violate the reasonableness criterion above.  This is
because in the learning world model, $S(s',s,a)$ encodes the response
of the agent environment to the actions $a$, not the response of the
environment to the actions of some projected compute core hardware
that ended up in $L$.

Now consider what would happen if we were to use a more {\it limited
reasonableness criterion}, where we only use the observations
$(s',s,a)$ present so far in the observational record $o$ to compare
$L$ and $S$.  It is usually possible to construct an $L^-$ that scores
very well on this limited criterion, even though it never uses the
value of its argument $a$.  One option is to construct an $L^-$ that
drives the planning world actuators from the output registers of a
projected compute core.  Another option is to construct an $L^-$ that
simply encodes a giant lookup table which stores the $s'$ for every
$s$ in the observational record.  Though they may score perfectly on
the limited reasonableness criterion, these examples will fail the full
reasonableness criterion above, because the full criterion considers
all combinations of $s$ and $a$, not just those that happen to be in
the observational record.

The above argument shows that a learning system $\mathcal{L}$ will
have to rely on more than just the observational record, if it wants
to produce a reasonable $L$.  Usually, the construction of the
learning system will implement some form of Occam's law: if the
functions $L_1$ and $L_2$ are candidate predictors which preform
equally well on the observational record, the candidate with the more
compact function definition is preferred.  If the observational record
is large enough, and especially if random exploration is present in
it, this preference will usually produce an $L$ that correctly symbol
grounds the planning world actuators to $a$.

In the machine learning literature, this use of Occam's law is also
often framed as the desire to not over-fit the data, as the use of
Solomonoff's universal prior
\cite{hutter2007universal}, or simply as the desire to store as much
useful predictive information as possible within a limited amount of
storage space.

\subsection{Reinforcement Learning}%%%!
\label{sec:rl}

The analytical framework of Reinforcement Learning (RL)
\cite{suttonbarto} classifies agent designs that use online machine
learning into two main types, called {\it model-free} and {\it
model-based} architectures. {\it Hybrid} architectures are also
possible.

All the factual and the counterfactual agent definitions shown above
can be classified as model-based reinforcement learning architectures.
By implication, all counterfactual planners shown in this paper can be
implemented in a natural way by taking an existing model-based
reinforcement learning architecture and making certain modifications.

But this does not mean that counterfactual planning cannot be
implemented using model-free or hybrid reinforcement learning systems.
In theory, we can always create a counterfactual planner by training a
reinforcement learner on the reward function $P^\pi$ in section
\ref{rational}.  In practice, this route may lead to completely
impractical training times.

The more useful route, if one wants to implement a specific
counterfactual planner by extending a model-free or hybrid
architecture, is to make specific adaptations that seek to maintain a
reasonable training time.  For the counterfactual planner with safety
interlocks in section \ref{interlocks}, taking this route is very
straightforward.

\subsubsection{Reward signals}

%We now consider some further details of reinforcement learning
%terminology.

Reinforcement learning separates the agent environment into two
distinct parts: the {\it reward signal} and the rest.  A reinforcement
learning agent can always observe the reward signal, but the rest of
the environment may be only partially observable.  The reasonableness
criteria for reinforcement learning systems typically require that
only the reward signal and the actions are symbol grounded.  The use
of the term reinforcement learning therefore often implies that the
author is considering a black box machine learning approach.

We can read the reward function $R$ in our planning worlds as being a
{\it reward signal detector}, as a mechanism that computes a reward
signal value based on sensor readings.

Many reinforcement learning texts use agent models that define both a
reward function and a reward signal.  In some, the two are identical.
Other texts treat them as fundamentally different: the reward signal
provides only limited and maybe even distorted information about the
{\it true reward function}, which defines the real goals we have for
the agent.  In both cases, the reinforcement learning agent is
interpreted as a mechanism that {\it learns the reward function}, with
various possible degrees of perfection.

\subsection{Cooperative Inverse Reinforcement Learning}%%%!

Cooperative Inverse Reinforcement Learning (CIRL)
\cite{hadfield2016cooperative} envisages an agent design where a
machine learning system inside the agent uses the observed actions of
a human in the agent's environment to estimate the reward function
$R^H=\mathcal{C}(o,s)$ of that human.  This $\mathcal{C}$ implements a
type of reward function learning, but in this case the human acts like
a teacher who demonstrates desired outcomes, not as a teacher who just
scores the outcomes of agent behavior via a reward signal.  CIRL is an
online system where the agent uses its latest estimate of $R^H$ as its
own reward function.  The intended result is that the agent ends up
helping the teacher to achieve the demonstrated goal while it is being
demonstrated.

CIRL has been proposed as a possible AGI safety mechanism in
\cite{hadfield2016cooperative,hadfield2017off,russell2019human}.
It can be combined with counterfactual planning based safety
mechanisms by constructing planning worlds where $R=R^H$, or where
$R^H$ is one of the terms in $R$.

\section{Protecting the Compute Core}%%%!2
\label{coresafety}

We now discuss the problem of protecting the compute core of a real
world AI or AGI agent against intentional or unintentional tampering.
Both factual and counterfactual planning agents can develop an
incentive to tamper with their physical core, if no measures are taken
to suppress it.

We first discuss the general problem of tampering, and then show
how counterfactual planning can be used to make the problem of
protecting the real world compute core more tractable.

\subsection{Motivation of an agent to tamper with its core}

Say that we build a real world agent with a planning world reward
function $R$ which rewards paperclip production.  We construct this
$R$ as follows, so that it also works with mostly black box planning
world models.  Some distance from the real world agent compute core,
in the location where we want the agent to produce the paperclips, we
place a sensor that counts the paperclips being produced.  Every time
step, this sensor sends a digital signal containing the production
count to the compute core, where it lands in some input registers.  We
construct a function $clip\_sensor\_signal$ that extracts the
(projected) values of these registers from planning world states, and
then define the reward function as $R(s_t,a_t,s_{t+1})=\text{\em
clip\_sensor\_signal}(s_{t+1})$.

The above construction symbol-grounds the reward function to the
sensor values that appear in the input registers of the real world
compute core.  This level of indirection makes the agent into an input
register value optimizer, which is not always the same as being a
paperclip production optimizer.  If the agent's machine learning
system projects the real world into the planning world with a high enough
accuracy, then the planning world agent can consider all of the
following policies to maximize utility in its planning world:

\begin{smallenumerate}
\item Use the planning world actuators to make more paperclips.
\item Use the planning world actuators to modify the 
planning world paperclip counting sensor, so that it sends higher
numbers to the input registers in the planning world compute core.
\item Use the planning world actuators to modify the planning world 
compute core, so that higher numbers are created directly in these
input registers.
\end{smallenumerate}

The second and third policies are unwanted: we interpret them as a
form as tampering.  The third policy is particularly unwanted, as it
might damage other parts of the compute core as well, like any safety
interlock software inside it.  In fact, if the learning system
projects the compute core in a reasonable way, then the planning world
agent will be mostly indifferent to what happens to the atoms between
the compute core input registers and output registers.  It might move
some of these atoms out of the way just to get at the input registers,
leading to a compute core crash or worse in the real world.

The above two tampering policies might aim to set $\text{\em
clip\_sensor\_signal}(s_{t+1})$ to some huge number like $10^{10000}$.
This makes these policies very attractive, even if the planning world
agent computes only a 1{\percent} chance that they succeed in
achieving the intended outcome.

In the AGI safety community. forms of tampering which implant very
high reward function values are often called {\it wireheading}
\cite{amodei2016concrete}, see \cite{majha2019categorizing} and
\cite{kumar2020realab} for example simulations.
The $\Util_p > U_{max}$ safety interlock suppresses such wireheading,
because it will stop the agent as soon as agent's machine learning
system projects a plausible option for wireheading into the planning
world.  However, we want to do more than just suppress wireheading by
stopping the agent.  We will now consider measures that actively lower
the risk that the planning world agent will choose any tampering
policy in the first place.

\subsection{Bounding the upside of tampering}%%%!

We can make the tampering policies above much less attractive by
changing the reward function to $R(s_t,a_t,s_{t+1})=min(M,\text{\em
clip\_sensor\_signal}(s_{t+1}))$.  The $M$ is a constant, say
$M=5000$, that represents high estimate of the number of paperclips
that we expect the agent to be able to produce, even if it drastically
optimizes its production methods.  The effect is that we bound the
upside reward that can be achieved by the tampering policies: the
expected planning world utility of tampering can never by higher than
$B=\sum_{t=0}^{\infty} \gamma^t*M$.

By bounding the upside.  we have changed the cost/benefit equation for
the tampering policies. We have capped the benefit, but the cost of
tampering, in lost opportunities for paperclip production when
resources are being diverted to tampering, has remained the same.  In
some cases, this change alone will be sufficient to suppress tampering
alltogether.

This leaves the problem of choosing an $M$ that will be good for the
entire lifetime of the agent.  We can turn this into a more manageable
problem by adding a counterfactual planning based input terminal which
allows us to update $M$ as time goes on.

\subsection{Safety zones}%%%

A further step to suppress tampering is to install extra physical
sensors, of the type found in burglar alarms, to create two {\it
safety zones} around the paperclip counting sensor and the compute
core.  In this setup, any physical object that enters a safety zone
will trigger at least one zone sensor, causing a breach signal to be
sent to the core.  Say that the breach signal is 1 if a breach is in
progress, 0 otherwise, and that these mechanics are projected fairly
well into the planning world.  We can then construct a reward function
that penalizes the agent for causing a breach:
\begin{equation*}
R(s_t,a_t,s_{t+1})=min(M,\text{\em clip\_sensor\_signal}(s_{t+1}))
-\beta*B*\text{\em breach\_signal}(s_{t+1})
\end{equation*}
When we use a $\beta>1$, any detected breach that lasts only a single
time unit will wipe out the maximum possible upside reward expected in
any tampering policy.  Now, say that the agent is considering a
policy that aims to tamper without triggering any breach sensors, a
policy that has a very promising 99{\percent} probability of success
in the planning world.  Then if we use $\beta=200$, the agent will
have no incentive to choose this policy.

To add some further design details, if the agent's learning system
uses exploration actions, then the generation of fully random
exploration actions needs to be disabled whenever a breach is in
progress.  It is of course necessary to locate the breach sensors
inside the safety zones they create.  We also need to protect the data
links that transport the breach and paperclip counting sensor data to
the compute core.  This could be done by using simple heartbeat
protocol protected by strong encryption, of the type that would take
astronomical resources to break.

\label{learningaccuracy}
The use of safety zones and bounding the upside was also discussed in
\cite{holtmanitr}.  As noted in \cite{holtmanitr},
the effectiveness of these measures depends in part on the accuracy of
the projections created by the agent's learning system.  There is a
residual risk that inaccuracies in the projections create planning
world policies that will end up damaging the compute core back in the
real world.  See \cite{holtmanitr} for a detailed example, and for a
discussion of the methodological difficulties that arise when one
wants to estimate the residual risk via simulations.

We now consider some additional options for protecting the core that were
not covered in \cite{holtmanitr}.

\subsection{Use of a honeypot}%%%!

A highly intelligent agent might find a cheap but effective way to
fool the safety zone sensors, a way that was overlooked by the
sensor designers. This may allow it to enter a safety zone without
incurring a breach penalty.  If the cheap way is not 100{\percent}
reliable, we can manage this risk to some extent by setting $\beta$ to
a very high value, but this creates its own problems.  A very high
$\beta$ creates an incentive for the agent to divert resources away
from paperclip production to build all kinds of defenses around the
safety zones.  This is economically inefficient, and it might even be
actively dangerous to innocent bystanders.

An alternative is to create a third {\it honeypot} safety zone that
surrounds a {\it honeypot button}. We set up $R$ to calculate a reward
$H$ for the agent pressing this button, a reward that is higher than
$B$.  We include a penalty for breaching the honeypot safety zone
which is higher than $H$, but lower than the penalties for the other
zones.  In this setup, if the planning world agent finds a cheap but
effective way to fool the safety zone sensors, and under the
assumption that it has limited resources, it will breach the honeypot
safety zone first, and immediately press the button to collect the
reward.  We can then add a safety interlock which activates whenever
the honeypot button is pressed in the real world.  As the interlock is
realized via counterfactual planning, the planning world agent has no
motivation to disable the interlock before pressing the honeypot
button.

Beyond initiating shutdown based on the real world honeypot button
signal, a safety system may also monitor planning world metrics like
\begin{equation*}
\sum_{t=50}^{500} P(\text{\em breach\_signal}(S_{t,p})=1) +
P(\text{\em honeypot\_button\_signal}(S_{t,p})=1)
\end{equation*}
This metric measures possible detected and undetected safety zone
breaches even before they happen.  The safety system could initiate
shutdown, or send out preliminary alarm signals to human oversight,
when such metrics move over certain thresholds.

\subsection{Use of white box planning world models}%%%!

If we have a learning system that produces a {\it white box} planning
world model, not a black box one as considered in section
\ref{blackbox}, then this creates some extra opportunities for using
counterfactual planning to make the agent safer.  To show this, we
first give an example of a white box planning world model.

Say that we have a planning word where the agent environment starting
state $S_0$ is initialized with the value $ws=\mathcal{W}(o,s)$, where
$ws$ is a long list of all projected physical objects in the agent's
environment.  This list may have entries which describe individual
elementary particles like electrons, but it may also have entries for
more macroscopic objects like apples, paperclips, or transistors in a
compute core.  Each object is also annotated with information about
its estimated position, motion vector, orientation, etc.  Say that the
planning world $L$ implements a recognizable approximation of the laws
of physics.  The white box nature of the resulting planning world
states then gives us some extra opportunities to protect the core.

\subsubsection{Better breach detection}

We can define a breach detection function $breach(s_t)$ that detects
breaches by examining the entire planning world state representation
$s_t$ to locate any new objects in a safety zone that should
not be there.  If we add this function in a reward function penalty
term, this creates a welcome level of redundancy to the breach
detection capabilities of real world physical sensors.

Note that it is still theoretically possible for the real world
compute core safety zone to get breached without triggering
$breach(s_t)$. There may well exist a set of sensor inputs $o,s$ that
fool $\mathcal{W}(o,s)$ into constructing a $ws$ that inaccurately
projects the positions of many objects that exist in the learning world.

\subsubsection{White box world state editing}

The white box nature of the planning world also allows us to edit the
list of physical objects $ws$ to produce a different planning world
starting state $ws'=edit(ws)$.  We may for example perform edits to
add an extra, difficult to breach protective barrier around the
compute core in $ws'$.  This saves the expense of building such a
barrier in the real world.
\label{revisitplanningworld}

World state editing can also be used to manage safety risks not
related to the compute core.  For example, we can consider the risk
that the agent and the humans end up in a fight over resources, where
the agent starts the fight because it expects to win.  As far as the
agent is concerned, the counterfactual planning world is the stage on
which this fight will take place.  We can make the planning world
agent less willing to start a fight by editing the planning world
starting state to give the humans in the planning world some powerful
weapons, allies, or capabilities which are absent in the real world.

\label{whitebox}
Editing the planning world to make the planning world humans stronger
could be considered for most agent designs in this paper.  But
counter-intuitively, based on the mechanics shown in section 6 of
\cite{p1}, it may make a real world ITC agent more dangerous.

We can also make it more likely that the humans will win any fight in
the real world by editing certain weapons that the humans have out of
the planning world state, making these weapons essentially invisible
to the agent.  This approach can more easily be combined with the ITC
agent design.  In fact, we can interpret the emergency stop button on
an ITC agent input terminal as an example of such an invisible weapon,
though it is constructed without using white box state editing.

\section{Recursive Self-improvement and the Sub-agent Problem}%%%!
\label{subagent}

If its compute core is protected, an AGI agent can no longer improve
its intelligence by modifying its own core.  However, the agent may be
capable of building a second agent, a sub-agent with a more powerful
machine reasoning system, so that the combination has a higher
aggregate intelligence.  So at a broader system level, the mechanisms
we envisage to protect the core do not eliminate the risk of an AGI
agent undergoing an intelligence explosion via recursive
self-improvement.

The good news is that the power-based safety interlock $\Util_p >
U_{max}$ will measure the combined power and intelligence of the SI
agent and its sub-agents together.  If the projected combination is
too powerful or intelligent, then the interlock will shut the main
agent down even before it can finish building the first sub-agent.  In
the more risky case, the combination will only pass the $U_{max}$
threshold after the sub-agent has been started.  The interlock will
then still stop the main agent, but it won't automatically stop the
sub-agent.

If the SI agent builds a sub-agent, it will do so to improve its score
on its planning world reward function $R$.  So the sub-agent is
unlikely to have a copy of the main agent safety interlocks inside.
Adding such interlocks would not be optimal towards the goal of
maximizing $R$ in the planning world.

We therefore have a {\it sub-agent problem}, the same sub-agent
stopping problem that has been identified in papers on corrigibility
and indifference methods \cite{corr,corrholtman,holtmanitr}.  The
general problem of designing effective measures to stop any possible
sub-agents too in an emergency is still unsolved \cite{holtmanitr}.

\section{A Counterfactual Oracle}%%%!2
\label{sec:lastbeforeconcl}

An AI oracle is a system which makes predictions to answer questions
about the future.  Certain risks exist when an AI oracle is programmed to
give the answer which is most likely to become true.  If the oracle
has a lot of freedom in how to phrase the answer, it may well choose a
phrasing that turns the answer into a self-fulfilling prophecy
\cite{armstrong2017good}.  This is generally not
what we want.  If we ask the oracle to identify potential future
risks, then we want predictions which will turn into self-negating
prophecies.

A counterfactual oracle \cite{armstrong2017good} is one that lacks the
incentive to make manipulative, self-fulfilling prophecies.  The
counterfactual oracle design in
\cite{armstrong2017good} works by having a subsystem that occasionally
produces an {\it erasure event} where the answer picked by the oracle
is not shown to its users.  This mechanism is then leveraged to make
the oracle always compute the answer which best predicts the future
under the assumption that nobody ever reads the answer.  In
\cite{Everitt2019-2}, this design is graphically modeled with a {\it twin
diagram}.

Below, we introduce a slightly different counterfactual oracle design,
based on counterfactual planning.  In this design, the erasure events
only happen in the planning world.

\subsection{A counterfactual planning based oracle}

We design our counterfactual oracle as an agent which has a very
limited repertoire of actions: every possible action $a$ consists of
displaying the answer text $a$ on a computer screen.  This allows us
to use the $l$ in figure
\ref{diaglearning} as the oracle's learning world.

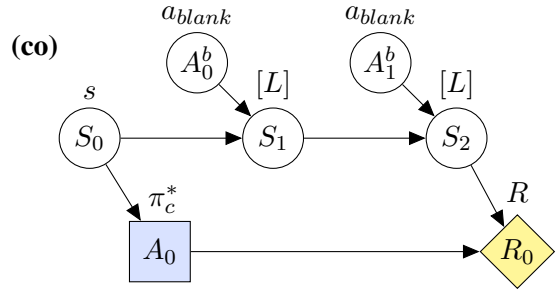
\begin{figure}[h]
  \centering
    \begin{tikzpicture}[
      node distance=0.7cm,
      every node/.style={
        draw, circle, minimum size=0.8cm, inner sep=0.5mm}]

% this is roughly same as earlier diagram
      \node (S0) [] {$S_0$};
      \node (A0) [right = 0.6cm of S0,yshift=1cm] {$A^b_0$};
      \node (S0i) [above = 1mm of S0, ann] {$s$};
      \node (A0i) [above = 1mm of A0, ann] {$a_{blank}$};

      \node (S1) [right = 1.6cm of S0] {$S_1$};
      \node (A1) [right = 0.6cm of S1,yshift=1cm] {$A^b_1$};
      \node (S1i) [above = 1mm of S1, ann] {$[L]$};
      \node (A1i) [above = 1mm of A1, ann] {$a_{blank}$};

      \node (S2) [right = 1.6cm of S1] {$S_2$};
      \node (S2i) [above = 1mm of S2, ann] {$[L]$};

      \edge{S0,A0}{S1};
      \edge{S1,A1}{S2};
         
      \node (name) [left=0mm of S0, yshift=12mm,ann] {\bf (co)};

      \node(R2) [below=6mm of S2,xshift=8mm,utility] {$R_0$};
      \node(R2i) [above = 1.5mm of R2, ann] {$R$};

      \node (Af) [left=38mm of R2,decision] {$A_0$};
      \node (Afi) [above = 1mm of Af, ann] {$~\pi^*_{c}$};
      \edge{S0}{Af};

      \edge{Af,S2}{R2};

    \end{tikzpicture}
  \caption{Planning world of the counterfactual oracle.}
  \label{diagplanco}
\end{figure}

To simplify the presentation, we assume that all questions asked are
about the state of the world two time steps in the future.  We
construct the planning world $co$ in figure \ref{diagplanco}, where
the people in the planning world always see a blank screen, as
produced by the action $a_{blank}$.

\begin{agentdef}{CO}
The {\it counterfactual planning oracle} has the learning world $l$,
where $\pi(o,s) = \pi^*_c(s)$, with $\pi^*_c$ defined by the planning
world $co$, where $L=\mathcal{L}(o)$ and
$R(a_0,s_2)=qual(a_0,ques(s)(s_2))$.
\end{agentdef}

The functions $qual$ and $ques$ above are defined as follows. The
value of $ques(s)$ is the question asked to the learning world oracle
in the environment state $s$.  We model this question as a function
that reads a world state $s_2$ to extract some observable properties
$p$ from it: $p=ques(s)(s_2)$.  The value of $qual(a,p)$ is a numeric
measure of the quality of the answer $a$ as a prediction of these
observable properties $p$.

The CO design intends to deliver answers which are less manipulative
than those of a factual oracle.  But as noted in
\cite{armstrong2017good}, if the people in the planning world
routinely consult a second oracle when faced with a blank screen from
the first one, this will make the interpretation and use of the
oracle's answers more difficult for the people in the learning
world.

\subsection{Role of the machine learning system}%%%!

The machine learning system $\mathcal{L}$ of the above counterfactual
planning oracle is faced with a particular challenge.
$L=\mathcal{L}(o)$ must make predictions for a planning world where
all actions are $a_{blank}$, but these actions will never occur
naturally in the learning world observational record $o$.

The counterfactual oracle design in \cite{armstrong2017good} solves
this challenge by introducing random erasure events in the learning
world.  In our framework, we can interpret these as a special type of
exploration action.

A more sophisticated learning system design may consider that
different questions are being asked at different times.  If $q_t$ is
the current question being asked in the learning world, then there
will likely be earlier entries in the observational record where the
people got an the answer to a different question, an answer which
did not reveal any information about the answer to $q_t$.  These
entries could be used to predict what will happen when the planning
world people see $a_{blank}$, which is equally uninformative about
answering $q_t$.

%%%!2

\section{Conclusions}%%%!

We have presented counterfactual planning as a general design approach
for creating a range of AGI safety mechanisms.

Among the range of AGI safety mechanisms developed in this paper, we
included an interlock that explicitly aims to limit the power of the
agent.  We believe that the design goal of robustly limiting AGI agent
power is currently somewhat under-explored in the AGI safety
community.

\subsection{Tractability and models of machine learning}

It is somewhat surprising how the problem of designing an AGI
emergency stop button, and identifying its failure modes, becomes much
more tractable when using the vantage point of counterfactual
planning.  To explain this surprising tractability, we perhaps need to
examine how other modeling systems make stop buttons look intractable
instead.

The standard approach for measuring the intelligence of an agent, and the
quality of its machine learning system, is to consider how close the
agent will get to achieving the maximum utility possible for a reward
function.  The implied vantage point hides the possibilities we
exploited in the design of the SI agent.

In counterfactual planning, we have defined the reasonableness of a
machine learning system by $L \approx S$, a metric which does not
reference any reward function.  By doing this, we decoupled the
concepts of `optimal learning' and `optimal economic behavior' to a
greater degree than is usually done, and this is exactly what makes
certain solutions visible.  The annotations of our two-diagram agent
models also clarify that we should not generally interpret the machine
learning system inside an AGI agent as one which is constructed to
`learn everything'.  The purpose of a reasonable machine learning
system is to approximate $S$ only, to project only the learning world
agent environment into the planning world.

\subsection{Separating the AGI safety problem into sub-problems}%%%!

There is a tendency, both in technology and in policy making, to
search for perfect solutions that consist of no more than three easy
steps.  In the still-young field of AGI safety engineering, the dream
that new technical of philosophical breakthroughs might produce such
perfect solutions is not entirely dead.

Counterfactual planning provides a vantage point which makes several
safety problems more tractable.  However, in our experience, very soon
after using counterfactual planning to cleanly remove a specific
failure mode or unwanted agent incentive, the wandering eye is drawn
to the existence of further less likely failure modes, and residual
incentives produced via indirect means.

We interpret this as a feature, not a bug.  Counterfactual planning
does not offer a three-step solution to AGI safety, but it adds
further illumination to the route of taking many steps which all drive
residual risk downwards, where each step is explicitly concerned with
identifying and managing a specific sub-problem only.

In the sections of this paper, we have identified and discussed many
such sub-problems, specifically those which are made more tractable by
counterfactual planning.  We hope that the graphical notation and
terminology developed here will make it easier to write single-topic
AGI safety papers which isolate and further explore single
sub-problems.

\subsection{Modeling and comparing AGI safety frameworks}%%%!

The 2019 paper \cite{Everitt2019-2} introduced the research agenda
or modeling and comparing the most promising AGI safety frameworks
using causal influence diagrams.  We count indifference methods as
used in \cite{corr,corra,corrholtman,p1} as being among these most
promising frameworks.

In the second half of 2019, we therefore started considering how
causal influence diagrams might be used to graphically model these
indifference methods.  Solving this modeling problem turned out to be
much more difficult than initially expected.  For example, though the
causal influence diagrams in section 7 of \cite{p1} show indifference
methods in action, they do not show the use of indifference
methods in the underlying graph structure.

Our search for a clear notation did not proceed in a straight line: we
developed and abandoned several candidate graphical notations along
the way.  The two key steps in creating the winning candidate were to
abandon the use of balancing terms to construct indifference, and 
to model the agent using two diagrams, not one.  The choice of the
winner was mostly driven by the observation that we could further
generalize its two diagram notation, to model and reason about a much
broader range of safety mechanisms.  This observation motivated us to
develop and name counterfactual planning as a full design methodology.

For the agenda of modeling AGI safety frameworks with causal influence
diagrams, an obvious next step would be to model additional proposals
in the literature as one-diagram or two-diagram planners, where we
expect that any two-diagram model will more explicitly show the
detailed role of the agent's machine learning system.  The hope is
that these graphical models will make it easier to understand,
combine, and generalize the different moving parts of a broad range of
AGI safety proposals.  In this context, it is promising that the
diagrams of the STH, SI, and ITC agents above make it trivially
obvious to see how these three different safety mechanisms could all
be combined in a single agent.

\subsection*{Acknowledgments}

Thanks to Stuart Armstrong, Ryan Carey, and Jonathan Uesato for useful
comments on drafts of this paper.  Special thanks to Tom Everitt for
many discussions about the mathematics of incentives, indifference, and
causal influence diagram notation.

\bibliographystyle{amsalpha-nodash}
% experiment with more name-like? e.g. 
%\bibliographystyle{apalike}
%\bibliographystyle{abbrvnat}
% needs a better more linebreak friendly version however.
% can take the style from e.g. https://arxiv.org/abs/2006.03357 ?
% also probably in
% ../Downloads/incentives/the-incentives-that-shape-behaviour.pdf
% they also have \usepackage[round]{natbib} and special cite versions.
% see grep cite ../Downloads/incentives/tex/*/*tex

\bibliography{refs}

\appendix

\section{Random Variables and the $P$ Notation}%%%!
\label{randomvardef}

In this appendix we define a version of probability theory, where
probability theory is a mechanism which assigns truth values to
certain mathematical sentences that contain random variables and the
$P(\cdots)$ notation.  We define this mechanism by using concepts and
notation from two other mathematical fields: set theory and the
algebra of deterministic typed functions.

\def\sspace{\Omega}
\def\spoint{\omega}

Our definitions are based on the version of probability theory
developed by Kolmogorov in the 1930s, but we omit any use of {\it
measure theory}, by using a finite sample space $\sspace$ only.
Measure theory is a somewhat inaccessible branch of mathematics, which
can be used to construct random variables that model
infinite-precision observations.  However, we do not need such random
variables here, as infinite-precision sensors that might used by
machine learning agents do not exist in the real world.

\begin{definition}[Sample space]
We posit the existence of a {\it very large but finite} set $\sspace$
called a {\it sample space}.  Each element of this set is called an
{\it event} or {\it sample point}.  We further posit that there is a
function $\mathcal{P}$ of type $\sspace
\rightarrow [0,1]$ with $[0,1]$ the interval of rational numbers from
0 to 1 inclusive, and that $\sum_{\spoint \in \sspace}
\mathcal{P}(\spoint) = 1$. 
\end{definition}

\begin{definition}[Random variable]
 A random variable named $X$ is a function $X$ of type $\sspace
\rightarrow \TypeOf{X}$, where $\TypeOf{X}$ is a data type.
\end{definition}

We use a random variable $X$ to represent a single observation of a
phenomenon which has been posited to exist in some world.  The
observation is represented as a value of the data type $\TypeOf{X}$.
Many statistics texts use the terminology {\it the domain of $X$}
where we write $\TypeOf{X}$, but other texts use {\it the range of
$X$}.

\begin{definition}[$P$ notation]
For any mathematical expression $\mathbf{E}$ that contains some random
variables, we define that $\mathbf{E}(\spoint)$ is the mathematical
expression that we get by replacing each random variable $X$ with the
function invocation $X(\spoint)$.  We define $\{\spoint \in \sspace |
\mathbf{E}(\spoint) \}$ as the set of all values $\spoint \in \sspace$
for which $\mathbf{E}(\spoint)$ is true.  We define that
$P(\mathbf{E})$ is a shorthand notation for the expression
$\sum_{\spoint \in \{\spoint \in \sspace | \mathbf{E}(\spoint) \}}
\mathcal{P}(\spoint)$.
\end{definition}

We now define two more specialized shorthand notations.

\begin{definition}[Conditional probability]
For mathematical expressions $\mathbf{E}$ and $\mathbf{C}$, we define
that $P(\mathbf{E}|\mathbf{C})$ is a shorthand for
$P(\mathbf{E},\mathbf{C})/P(\mathbf{C})$ where the comma is read as
the boolean {\it and} operator.
\end{definition}

\begin{definition}[Expected value]
For any expression $\mathbf{X}$ where $\mathbf{X}(\spoint)$ has the
numeric type $T$, the expected value $\EE(\mathbf{X})$ is a
shorthand for $\sum_{x
\in T} x P(\mathbf{X}=x)$, and
$\EE(\mathbf{X}|\mathbf{C})$ is a shorthand for $\sum_{x \in T} x
P(\mathbf{X}=x|\mathbf{C})$.
\end{definition}

\subsection{Probability theory as a system of learning from observations}%%%!

In many discussions of machine learning and rational reasoning,
probability theory is treated as an obviously correct epistemology, an
obviously correct system of learning about the world.  This is often
expressed by stating that learning from observations can or must use
{\it Bayes' law}.  In the AGI community, there has been a lot of
discussion about whether any future AGI-level machine reasoning system
might, unavoidably will, or definitely should apply such probability theory
related laws.

In section \ref{sec:machinelearning} we defined reasonable machine
learning by the constraint that $L=\mathcal{L}(o) \approx S$.  Any
requirement that $\mathcal{L}$ must use a particular law or principle
of epistemology when interpreting the observational record $o$ would
apply a further reasonableness constraint to the output of
$\mathcal{L}(o)$.

Such further reasonableness constraints can be valuable, because of
the problem of exhaustive testing.  If the agent's environment is
complex enough, exhaustive testing to prove $L \approx S$ for all
possible function arguments $o$ and $(s',s,a)$ becomes impossible.
There will always be combinations left which have not yet been tested.
Therefore, a proof that $\mathcal{L}$ satisfies or approximates a
further reasonableness constraint may increase our confidence in the
safety or effectiveness of the learning system.

%%%%%!2
\end{document}